\documentclass[10pt,twocolumn,letterpaper]{article}

\newcommand{\real}{\texttt{\textbf{RealPC}}}

\newcommand{\topodg}{\texttt{\textbf{TopODGNet}}}
\newcommand{\bosh}{\texttt{\textbf{BOSHNet}}}

\usepackage{simpleConference}
\usepackage{times}
\usepackage{graphicx}
\usepackage{amssymb}
\usepackage{epsfig}
\usepackage{amsmath}
\usepackage[pagebackref,breaklinks,colorlinks]{hyperref}
\usepackage{caption}
\usepackage{subcaption}
\usepackage{algorithm}
\usepackage{algpseudocode}
\usepackage{float}
\usepackage{xcolor}
\usepackage{booktabs}
\usepackage{amsfonts}
\begin{document}

\title{Revisiting Point Cloud Completion: Are We Ready For The Real-World?}

\author{Stuti Pathak\textsuperscript{\textdagger}\\
University of Antwerp\\
{\tt\small stuti.pathak@uantwerpen.be}
\and
Prashant Kumar\textsuperscript{\textdagger}\\
Indian Institute of Technology Delhi\\
{\tt\small prashantk.nan@gmail.com}
\and
Dheeraj Baiju\\
Birla Institute of Technology and Science Pilani\\
{\tt\small f20212232@pilani.bits-pilani.ac.in}
\and
Nicholus Mboga\\
Geographic Information Management\\
{\tt\small nicholus.mboga@gim.be}
\and
Gunther Steenackers\\
University of Antwerp\\
{\tt\small gunther.steenackers@uantwerpen.be}
\and
Rudi Penne\\
University of Antwerp\\
{\tt\small rudi.penne@uantwerpen.be}
}

\twocolumn[{%
\renewcommand\twocolumn[1][]{#1}%
\maketitle
\begin{center}
\setlength{\belowcaptionskip}{0pt}
    \centering
    \captionsetup{type=figure}
    \includegraphics[width=\textwidth,  trim={1cm, 50cm, 21cm, 1.5cm}, clip]{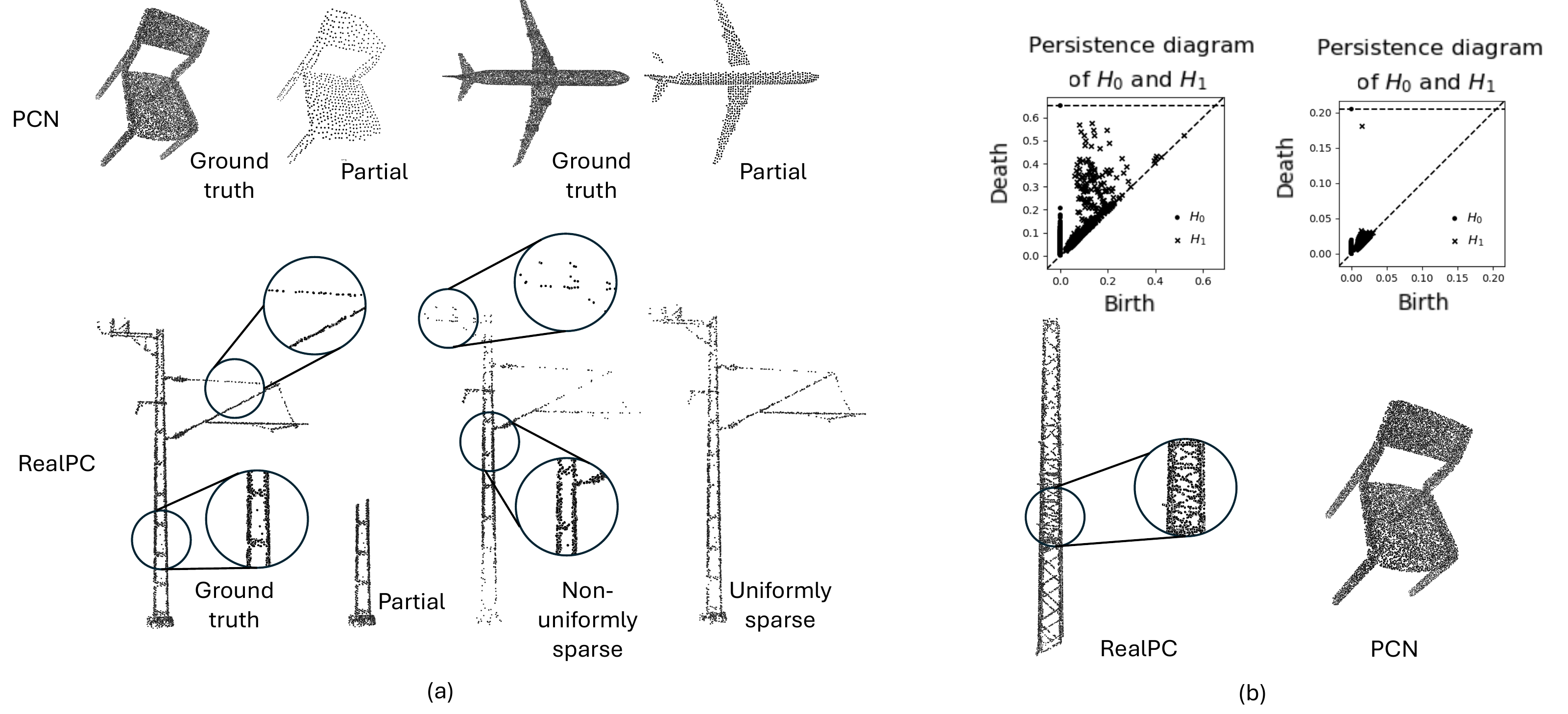}
    \caption{Qualitative comparison of current existing and \real{} object point clouds. (a) We show a complete absence of non-uniformity and noise for the PCN dataset \cite{yuan2018pcn}, in contrast to multiple levels of non-uniformity (as shown by the magnified regions) along with noise (bottom two magnifications) in the case of our \real{} dataset. (b) Comparison of Persistence Diagrams of a point cloud each from \real{} and PCN dataset. Points further away from the diagonal indicate strong topological features. Unlike PCN, \real{} has numerous significant 0- and 1-dimensional (\textit{$H_0$, $H_1$})  topological features.}
\label{fig:nonuni_topo}
\end{center}%
}]

\begin{abstract}
\renewcommand{\thefootnote}{}
\footnotetext{\textsuperscript{\textdagger}Equal contribution.}

\textit{Point clouds acquired in constrained, challenging, uncontrolled, and multi-sensor real-world settings are noisy, incomplete, and non-uniformly sparse. This presents acute challenges for the vital task of point cloud completion. Using tools from Algebraic Topology and Persistent Homology ($\mathcal{PH}$), we demonstrate that current benchmark object point clouds lack rich topological features that are integral part of point clouds captured in realistic environments. To facilitate research in this direction, we contribute the first real-world industrial dataset for point cloud completion, \real{} - a diverse, rich and varied set of point clouds. It consists of $\sim$ 40,000 pairs across 21 categories of industrial structures in railway establishments. Benchmark results on several strong baselines reveal that existing methods fail in real-world scenarios. We discover a striking observation - unlike current datasets, \real{} consists of multiple 0- and 1-dimensional $\mathcal{PH}$-based topological features. We prove that integrating these topological priors into existing works helps improve completion. We present how 0-dimensional $\mathcal{PH}$ priors extract the global topology of a complete shape in the form of a 3D skeleton and assist a model in generating topologically consistent complete shapes. Since computing Homology is expensive, we present a simple, yet effective Homology Sampler guided network, \bosh{} that bypasses the Homology computation by sampling proxy backbones akin to 0-dim $\mathcal{PH}$. These backbones provide similar benefits of 0-dim $\mathcal{PH}$ right from the start of the training, unlike similar methods where accurate backbones are obtained only during later phases of the training.}

\end{abstract}

\section{Introduction}
\label{sec:intro}
In its simplest form, a \textit{point cloud} (PC) is a discrete surface-sampling of an object or an environment. With the growing interest in digital twins of the real-world, PCs can now be obtained using various sensors and techniques, such as LiDAR, depth cameras, photogrammetry, structured light scanning, etc. Their ability to represent complex shapes with attention to local intricate details makes them indispensable across numerous fields, such as robotics \cite{geng2023partmanip}, autonomous driving \cite{yuan2024ad}, medical image analysis \cite{zhao2023pointneuron}, augmented and virtual reality \cite{nguyen2023impact}, remote sensing and geoinformatics \cite{guo2023kd}, etc.


Inherently, real-world PCs are non-uniformly sparse, noisy, complex-structured and topologically rich, unlike their synthetic counterparts (Figure \ref{fig:nonuni_topo}). Moreover, they may also have missing parts due to occlusions, view-angle constraints, object surface properties, environmental factors, etc. This hinders the performance of all downstream PC tasks, for instance, registration, surface reconstruction, object recognition, segmentation, etc. As a result, obtaining the full 3D shape representation of an incomplete and sparse PC plays a vital role in the practical applications of PC datasets. With the rise of data-driven methods, this area has been widely investigated under the umbrella term \textit{point cloud completion} \cite{fei2022comprehensive, zhuang2024survey, tesema2023point}.



Although historically motivated from a real-world perspective, as discussed above, most of the existing models in the field of machine learning-based PC completion report their performances on synthetic datasets \cite{Wang_Cui_Guo_Li_Liu_Shen_2024, zhang2023learning, chen2023anchorformer, li2023proxyformer, xiang2021snowflakenet, zhou2022seedformer}. This stems from the easy accessibility, simpler shapes, minimal to no noise, and uniform point distribution of controlled and synthetically-generated PC datasets \cite{yuan2018pcn,tchapmi2019topnet,pointtr,vrcnet}. However, due to the excessive tailoring of these models to these existing datasets, they fall short of providing similar results for real-world datasets.


The unordered representation of a PC itself does not allow for the encoding of any structural information, which makes PC processing very challenging. \textit{Persistent Homology} ($\mathcal{PH}$), a powerful tool from \textit{Topological Data Analysis} (TDA), has proven to be highly effective in extracting structural properties from 3D datasets.
It has gained traction in vision applications, particularly for analyzing and understanding the underlying shape and structure of data from different modalities such as PCs, images, graphs \cite{singh2023topological,nguyen2020bot,cang2018representability}, etc. However, as far as our knowledge goes this line of research has not yet been explored for the task of completing PCs in general, let alone real-world ones. Hence, this paper presents the following key contributions:


\begin{itemize}
\item We present a real-world paired (ground truth complete and sparse/partial incomplete) industrial object PC dataset \real{}, captured in challenging, uncontrolled and realistic multi-sensor settings.

\item We demonstrate that \real{} PCs are intrinsically different from existing synthetic ones via three important metrics, $\mathcal{PH}$ features, non-uniformity and noise. Using tools from TDA, we discover that they have rich and versatile topological features. Benchmarking exercises on non-neural and neural methods reinforce our discovery and establish that current methods for existing datasets do not work well with such realistic ones.

\item We prove the significance of integrating 0-dim $\mathcal{PH}$ priors into existing completion models (under certain conditions). Since Homology extraction is costly, we design a simple and intuitive \textit{Homology Sampler} network \bosh{}, which extracts multiple Homology backbones from a PC at various scales. These priors force a network from the start of the training to stick to the backbone while generating complete shapes, enabling precise completion.

\end{itemize}

\section{Related Work}
\label{sec:rw}

\subsection{Benchmark Object-Level Point Cloud Datasets}
\label{sec:syn_real}

Not only is the processing of real-world PC data arduous but also its acquisition comes with its own set of challenges. With the availability of high-end 3D-scanning devices nowadays, this task ultimately boils down to balancing the time involved, the accuracy required, and the setup costs. In addition, there is no particular standard storage format followed by the available sensors in the market today. Moreover, such PC datasets have background outliers which can make most of the modern data-driven algorithms blow up and therefore demand manual outlier removal, which is extremely time-consuming. All of these factors combined pose a great difficulty in accessing and working with realistic PC data and introduce an inevitable hurdle for the 3D vision research community. Hence, researchers prefer working on synthetically-generated datasets instead.

Among some of the well-known datasets for PC completion is the PCN dataset \cite{yuan2018pcn}, which contains paired partial and complete PCs derived from 30,974 CAD models across 8 categories from the ShapeNet repository \cite{chang2015shapenet}. Two more datasets derived from the same CAD repository \cite{chang2015shapenet} are ShapeNet55 and ShapeNet34 datasets \cite{pointtr}. ShapeNet-55 includes 52,470 objects from 55 categories. The ShapeNet-34 dataset is derived by splitting the original ShapeNet-55 dataset into 34 seen categories and 21 unseen categories. Another such dataset is the MVP dataset \cite{vrcnet}, which consists of more than 100,000 PCs from 16 categories sampled again from CAD models. 

Although fewer in number, some recent real-world PC object datasets include OmniObject3D \cite{wu2023omniobject3d}, MatterPort3D \cite{chang2017matterport3d}, ScanNet \cite{dai2017scannet}, and KITTI \cite{geiger2012we}. The first three datasets offer PCs captured in extremely controlled environments, in contrast to real-world settings, which induce numerous factors of variation, disturbance and noise. Moreover, all of these datasets are unpaired and hence contain only partial shapes without ground truth, making them infeasible for supervised training.

\subsection{Point Cloud Completion}

PC completion has been widely explored for existing datasets such as PCN and ShapeNet. Most approaches frame this as learning point-wise features and generating points individually to reconstruct the complete structure. Voxel and point-based techniques have been used extensively for this purpose. GRNet \cite{xie2020grnet} uses 3D CNNs and MLPs for PC representation. Voxel-based methods struggle with limited input points, resulting in incomplete neighborhoods and hole-ridden outputs. Point-based methods operate directly on raw PCs (e.g. PCN \cite{yuan2018pcn}). Multiple works \cite{xiang2021snowflakenet,PMPNet,zhou2022seedformer} have used this strategy for effective completion. ODGNet \cite{ODGNet} incorporates a seed generation U-Net for enhanced seed representation, and orthogonal dictionaries to learn shape priors. PointAttN \cite{Wang_Cui_Guo_Li_Liu_Shen_2024} leverages cross and self-attention mechanisms, using Geometric Details Perception and Self-Feature Augment blocks to establish structural relationships between points. PointTr \cite{pointtr} represents PCs as unordered groups with position embeddings, converting them to point proxy sequences for generation via a transformer encoder-decoder architecture. There are several similar works \cite{GAN-RL,pcn,topnet,vrcnet,snowflakenet,PMPNet,RFNet,multi-view,mepcn,Vipc,alliegro2021denoise}, which revolve around the point feature learning approach.
The current literature works well on these PCs, but their performance on real-world PCs remains relatively less explored.

\subsection{Persistent Homology and Topological Deep Learning}

Persistent Homology ($\mathcal{PH}$), having its origins in algebraic topology, measures topological features, which characterize the shape of the data. These features help in discerning essential structures and patterns, which conventional methods cannot. Topological methods have been recently used across several learning-based tasks to augment architectures with better feature descriptors \cite{giansiracusa2017persistent,
pun2018persistent,hofer2017deep,clough2019explicit,liu2016applying,chen2019topological,dey2010persistent}. Vectorized topological descriptors extracted using $\mathcal{PH}$ have been used for interpretability of models and adversarial learning \cite{gabrielsson2019exposition}. \cite{gebhart2019adversarial} introduces topological representation for auto-encoding and demonstrates the topological richness of the learned representations. $\mathcal{PH}$ has also been used for several geometric problems like surface reconstruction, pose matching \cite{carlsson2004persistence,bruel2020topology,dey2010persistent}, etc. \cite{gabrielsson2020topology} proposes general-purpose differentiable topology layers to extract descriptors for topological regularization. 

\section{Background}
\label{sec:background}



Let $X$ be some general bounded surface in 3-space, discretely sampled to form a point cloud, represented by the unordered set
$\phi$. A standard tool from algebraic topology represents the topology of $X$ by a simplicial complex ${\cal C}$, a finite set of simplices of different dimensions. In our case the surface $X$ has dimension 2, so the only simplices we need are vertices (the points $\phi$), edges (connecting some pairs of $\phi$) and triangles (bounded by some triples of edges). By definition, the different simplices of this complex are either disjoint or intersect in a common subsimplex. The idea is that ${\cal C}$ provides a triangulation of $X$ such that the union of all simplices is topologically equivalent (homeomorphic) to $X$. The nice thing about representing a space $X$ by a simplicial complex ${\cal C}$ is that we can consider integer linear combinations of simplices at each dimensional level, yielding chains or cycles of edges and triangles, including so-called boundary maps
to obtain cycles of one dimension less. The benefit of this added algebraic structure is the {\em homology\/} of the complex ${\cal C}$, identifying vertices that are linked by edges, and detecting cycles
of edges that bound holes of the surface, yielding topological features of $X$, independent from the
chosen triangulation ${\cal C}$.


\begin{figure}
    \centering
\includegraphics[width=0.5\textwidth, trim={0cm, 65cm, 35cm, 0cm}, clip]{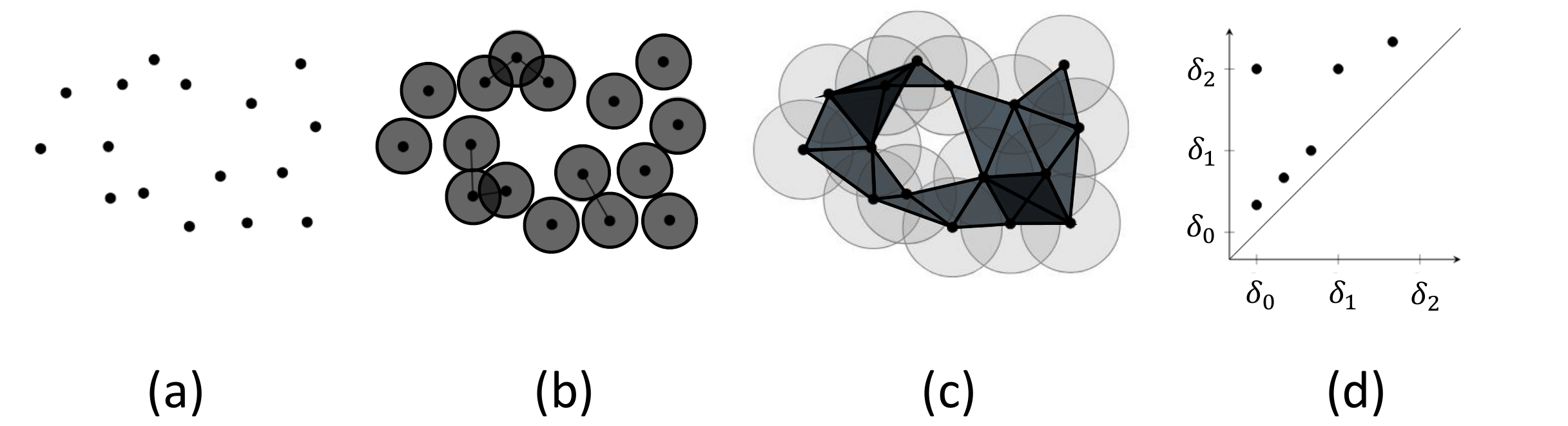}
\caption{(a) to (c) Progression of filtration on a PC over different spatial resolutions as the distance threshold increases \cite{moor2020topological}. (d) Birth and death of $k$-dim topological features documented in the form of a persistence diagram, i.e., $(b_i,d_i)$ pairs, so that each point corresponds to a homology which is born at $b_i$ and dies at $d_i$.}
\vspace{-0.68cm}
\label{fig:filtration_main}
\end{figure}

The idea of Persistent Homology ($\mathcal{PH}$) is to develop the homology of a surface in an incremental way,
starting from the sampled PC \cite{books/daglib/0025666,poulenard2018topological,rbgsurface,edelsbrunner2008computational}.
For a PC $\phi$, a finite sequence of subcomplexes of $\mathcal{C}$ is constructed, such that $\phi = \mathcal{C}_0 \subseteq \mathcal{C}_1 \subseteq \mathcal{C}_2 \subseteq \mathcal{C}_3....\mathcal{C}_i....\mathcal{C}_n = \mathcal{C}$. 
More precisely,
in Figure \ref{fig:filtration_main} we illustrate the development of such a sequence of subcomplexes
by increasing the radius of the $\alpha$-neighborhood balls around each point. Two points are connected by an edge when their balls intersect, i.e., distance between them $\leq$ 2$\alpha$. As $\alpha$ increases, we introduce more and more edges, following a specific order, implying an extension of our complex. Observe that these extensions only happen at a finite number of critical values for $\alpha$.
Consequently we obtain a finite sequence of increasing complexes. When an edge is added, it can potentially create a triangle. In this procedure, lower-dimensional simplices are always added before higher-dimensional ones. During the growing of the subcomplexes, we observe the creation and destruction of the topological features, as described by the homology. 
Indeed, when an edge is added, it may create a new hole that was not present before. This marks the birth of a feature. When an edge is added, it may fill an existing hole completely. This leads to the death of the feature that was born when the hole first appeared. For instance, consider four points forming a rectangle, hence creating a 1-dimensional hole (a cycle). When the diagonal edge is added later, it fills the rectangle with two triangles, causing the 1-dimensional hole to disappear.
Each 1-dimensional hole has a specific birth value of $\alpha$ and a death value of $\alpha$. The addition of an edge always results in either the creation or destruction of a homology.
In the case of a PC or an image, we don't know the optimal values of $\alpha$ to extract significant features. Therefore, we consider all possible values of $\alpha$ and track the changes in homology. This creates a nested sequence of simplicial complexes, known as a filtration. Each hole has a birth-death pair $(b,d)$, representing its persistence. These pairs can be represented as points in a 2D graph, called a Persistence Diagram, where the diagonal $(b=d)$ represents trivial pairs while points above the diagonal represent topological descriptors.


\section{\real{} Dataset}

\subsection{Creation}
\label{sec:cat_data}

In light of the challenges outlined in Section \ref{sec:syn_real} concerning the acquisition of real-world PC datasets, in this section we demonstrate a methodology (Figure \ref{fig:methodology}) to extract and process paired object PC datasets from existing open-source scene-level datasets. Their paired nature ensures suitability for any supervised PC completion algorithm, therefore, lowering the threshold for experimenting on real-world data. Although we have explored this methodology in a railway-environment setup, this exact procedure can be followed for any scene-level PC dataset.

We use four open-source scene-level annotated railway datasets from different countries, a few of which are visualized in Figure \ref{fig:scene_vis} (more about them in supplementary material). We extract annotated industrial structures from these scene-level PCs, hence obtaining a PC with multiple industrial structure shapes (Figure \ref{fig:methodology} Input to (A)). Our creation pipeline (Figure \ref{fig:methodology}) is divided into five parts. (A) We deploy HDBSCAN \cite{campello2013density}, a hierarchical clustering algorithm to cluster individual structures. (B) We perform manual inspection to extract complete industrial structures, which can serve as good ground truth (GT). Steps (C), (D), and (E) process these GT PCs in three different ways to induce uniform sparsity, non-uniform sparsity and incompleteness: (C) pick a random viewpoint around a GT PC and remove N points farthest away from this viewpoint, (D) pick a random viewpoint around a GT PC, and sample N points \textit{w.r.t} the probability values assigned to the points, which are  either proportional or inversely proportional to a point’s cubed distance from this viewpoint, and (E) randomly sample N points from a GT PC. While (C) and (D) are repeated for multiple viewpoints and N values, (E) is performed for several values of N. We manually examine all the industrial structures and classify them into 21 classes based on their geometric shapes, hence obtaining a real-world PC dataset which we call \real{}.

We shall make \real{} as well as the code for the processing pipeline open-source.


\begin{figure}
    \centering
    \includegraphics[width=0.5\textwidth, trim={0cm, 4cm, 0cm, 0cm}, clip]{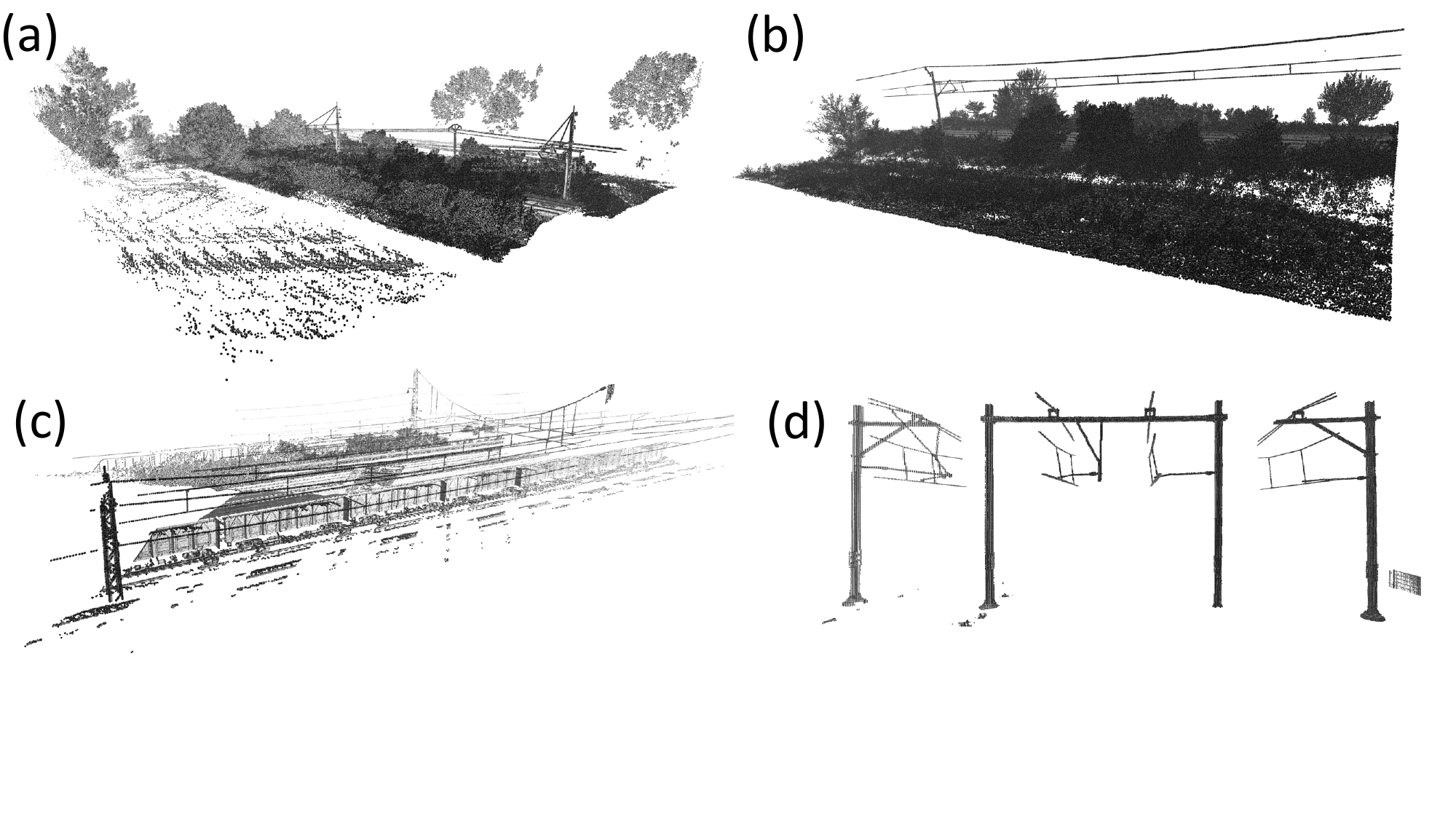}
    \caption{Scene-level PCs from different acquisition techniques (a) \cite{sncfNuagePoints}, (b) \cite{cserep2022hungarian}, (c) \cite{qiu2024whu}, and (d) \cite{ton2022labelled}.}
    \label{fig:scene_vis}
\end{figure}

\begin{figure}
    \centering
    \includegraphics[width=0.5\textwidth, trim={0cm, 55cm, 22cm, 0cm}, clip]{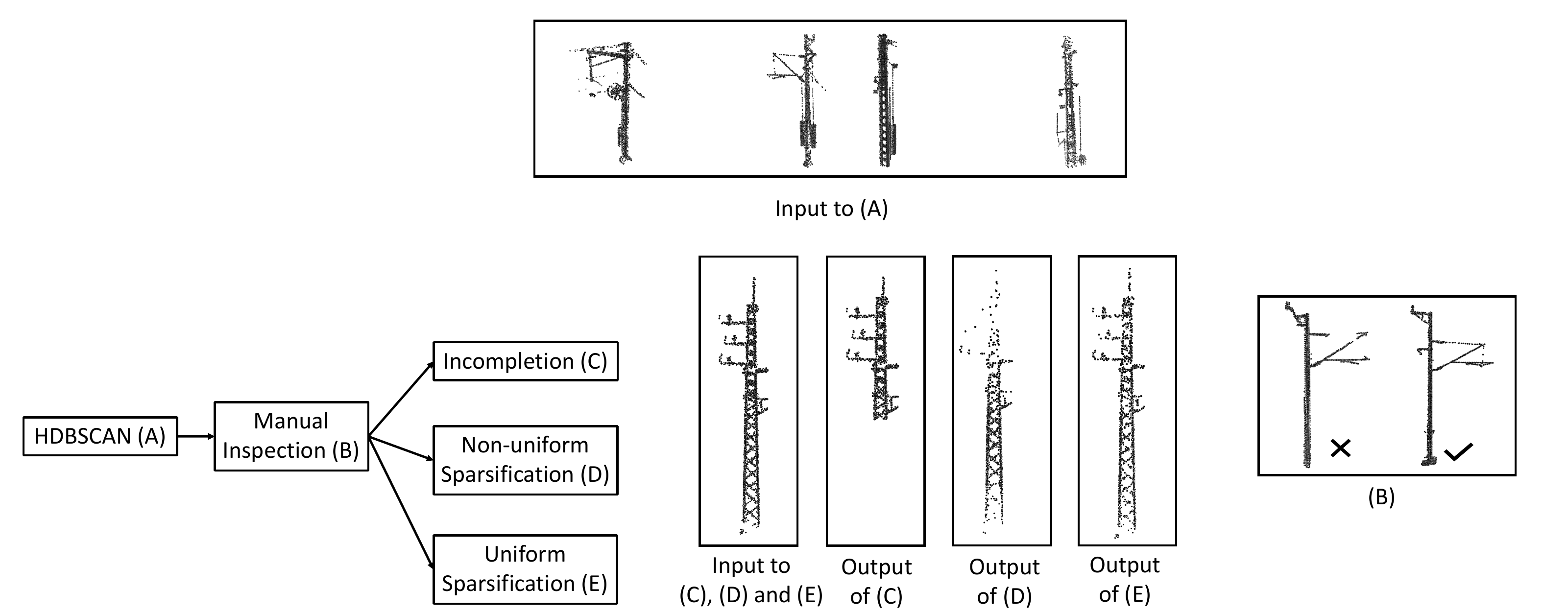}
    \caption{Object-level training dataset creation methodology. Input to (A) shows the segmented industrial structures from scene-level PCs. Figure (B) shows how manual inspection was done to extract ground truth data. Output of (C), (D), and (E) are the three variants of sparse and incomplete PCs.}
    \label{fig:methodology}
\end{figure}

\begin{table*}

    \centering
    \resizebox{\linewidth}{!}{
    \begin{tabular}{c@{\hskip 0.5cm}ccccc@{\hskip 0.5cm}ccccc@{\hskip 0.5cm}ccccc}
    \toprule
    {} & \multicolumn{5}{c}{Noise} & \multicolumn{5}{c}{Non-Uniformity} & \multicolumn{5}{c}{PH-based ($H_0$; $H_1$) }\\
    \cmidrule(r{0.5cm}){2-6} \cmidrule(r{0.5cm}){7-11} \cmidrule(lr){12-16} 
    {} & Class 1 & Class 2 & Class 3 & Class 4 & Mean & Class 1 & Class 2 & Class 3 & Class 4 & Mean & Class 1 & Class 2 & Class 3 & Class 4 & Mean\\
    \midrule
     PCN & 10.6 & 8.8 & 8.6 & 14.0 & 12.2 & 18.1 & 12.0 & 14.7 & 20.4 & 19.7 & 79.5; 34.2 & 52.4; 22.2 & 66.0; 31.9 & 87.1; 36.9 & 86.0; 37.5\\
     ShapeNet & 9.8 & 12.2 & 14.9 & 11.2 & 23.7 & 13.9 & 14.6 & 20.2 & 17.8 & 31.8 & 44.4; 18.6 & 47.2; 20.2 & 66.0; 29.4 & 56.4; 26.5 & 101.3; 41.0\\
     \real{} & \textbf{137.4} & \textbf{99.2} & \textbf{93.1} & \textbf{92.5} & \textbf{113.7} & \textbf{234.2} & \textbf{222.4} & \textbf{128.1} & \textbf{148.2} & \textbf{173.8} & \textbf{382.3}; \textbf{155.7} & \textbf{322.5}; \textbf{114.9} & \textbf{341.2}; \textbf{149.5} & \textbf{274.8}; \textbf{126.4} & \textbf{345.2; 155.5} \\
    \bottomrule
    \\
    \end{tabular}
    }
    \vspace{-0.5cm}
        \caption{Quantitative comparative analysis of \real{} against PCN and ShapeNet with respect to Noise ($\times 10^{-4}$), Non-uniformity ($\times 10^{-4}$), and 0 and 1-dimensional persistence of topological features ($\times 10^{-4}$). All numbers demonstrate that \real{} has stronger homology-based topological features, and higher non-uniformity and noise. Refer to Figure \ref{fig:nonuni_topo} for a qualitative comparison.}
    \label{tab:comparision}
\end{table*}

\subsection{Comparative Analysis}
\label{sec:compar_analysis}
We conduct a comprehensive geometric and topological analysis and comparison of \real{} with PCN and ShapeNet based on the following three factors:

\begin{itemize}
\item Persistent Homology ($H_0$; $H_1$): Persistence Diagrams in $\mathcal{PH}$ represent the $(birth, death)$ of topological features across a filtration, as points on the \textit{x-y} plane. All points are above the diagonal $(x=y)$. Points further from the diagonal represent features with larger persistence ($death-birth$), and hence are more important. We compute the average of persistence averages, for 0- and 1-dimensional topological features, across all classes of a given dataset and report them as \textit{PH-based ($H_0$; $H_1$)}.
    \item Non-Uniformity: For a PC we calculate the distance of all of its N points with their nearest neighbor and then find the standard deviation of all these distances across the entire PC. We report the average of all these standard deviation values across different classes of different datasets as \textit{Non-Uniformity}.
    \item Noise: For each point in a PC we first fit a plane using linear regression over its k-nearest neighbors. Then we calculate the perpendicular distance from the said point to the said plane. We average this distance for all the points in our PC. Finally, we calculate the average of these average values, termed \textit{Noise}, across different classes of different datasets.
\end{itemize}

For all three mentioned datasets, we present in Table \ref{tab:comparision} the above three metrics over four random classes and also the mean of \textit{all} the classes (not just four). Our dataset shows the highest \textit{Noise}, \textit{Non-uniformity}, and \textit{PH-based ($H_0$; $H_1$)} values consistently when compared to existing datasets. This result is supported by the visualizations shown in Figure \ref{fig:nonuni_topo} (a) and (b). We can clearly see high non-uniformity and noise in PCs from our dataset, in contrast to the complete absence of them in PCN. The (b) part of this figure investigates the existence of topological features in \real{} and existing PCs. The contrast between the two is striking. While the persistence diagram for ours consists of numerous important 0- and 1-dimensional topological features, existing datasets (e.g. PCN) lack them. This is indicated by the presence of several points further away from the diagonal for \real{}, and their absence for PCN.

\subsection{Benchmarking on Non-Neural Methods}

We evaluate and compare \real{} and ShapeNet on three critical and fundamental non-neural network based tasks for PCs: (a) Simplification using Weighted Locally
Optimal Projection (WLOP) \cite{huang2009consolidation} by $\times \; 0.3$, (b) Surface reconstruction using alpha shapes \cite{edelsbrunner1983shape} with $\alpha = 0.05, 0.1 \;\&\; 0.15$, and (c) Upsampling using \cite{huang2013edge} by $\times \;2$. We use Chamfer Distance and Hausdorff Distance as metrics, and average them across all the classes per dataset. We also average across multiple values of $\alpha$ for the task (b). For more details on these methods and metrics, please refer to the supplementary material.

As shown in Table \ref{tab:non_neural_tasks}, \real{} exhibits significantly higher errors when compared to ShapeNet for the three tasks, demonstrating the complexity of \real{}. For a visual comparison, please refer to the supplementary material.

\begin{table}
\small
    \centering
    \small 
    \setlength{\tabcolsep}{3pt} 
    \begin{tabular}{c@{\hskip 0.3cm}ccc}
    \toprule
    {} & Simplification & Surface Reconstruction & Upsampling \\
    {} & CD & HD & CD \\
    \cmidrule(){2-4}
    ShapeNet & 0.522 &  68.198 & 0.043\\
    \real{} & \textbf{8.817} & \textbf{494.591} &  \textbf{0.838}\\
    \bottomrule
    \end{tabular}
\vspace{-0.3cm}\caption{ Comparison of \real{} with ShapeNet on three popular non-neural tasks. We compare simplified/meshed/upsampled PCs with original PCs averaged across all classes. CD is Chamfer Distance and HD is Hausdorff Distance. All numbers are ($ \times 10^{-3}$).}
\label{tab:non_neural_tasks}
\end{table}



\begin{table}
\centering
\small
\setlength{\tabcolsep}{1pt}
\begin{minipage}{0.28\linewidth}
\centering
\begin{tabular}{c@{\hskip 0.01cm}c@{\hskip 0.15cm}c}
\toprule
{} & ShapeNet & \real{} \\
\cmidrule(lr){2-3}
CD  & 67 & \textbf{90} \\
EMD & 61 & \textbf{90} \\
\bottomrule
\end{tabular}
\end{minipage}
\hfill
\begin{minipage}{0.54\linewidth}
\centering
\begin{tabular}{c@{\hskip 0.01cm}c@{\hskip 0.15cm}c}
\toprule
{} & ShapeNet & \real{} \\
\cmidrule(lr){2-3}
FoldingNet  & 5 & \textbf{27} \\
DGCNN       & 49 & \textbf{58} \\
SnowflakeNet & 19 & \textbf{23} \\
\bottomrule
\end{tabular}
\end{minipage}
\vspace{-0.3cm}
\caption{\textbf{Left Table}: Benchmarking result for shape generation on diffusion-guided shape generation \cite{vahdat2022lion} using the 1-NNA metric (with both Chamfer distance (CD) and Earth Mover Distance (EMD)) as our main metric.
It quantifies the distributional similarity between generated shapes and
validation sets and measures both quality and diversity. \textbf{Right Table}: Benchmarking result for reconstruction using standard PC backbones using the Chamfer Distance metric. Both tables use Chamfer Distance  ($ \times 10^{-3}$).}
\label{tab:benchmark_reconstruction_gen}

\end{table}

\subsection{Benchmarking on Neural Methods}

\begin{table*}
    \centering
    \resizebox{\linewidth}{!}{
    \begin{tabular}{c@{\hskip 0.5cm}ccccccccccccccccccccccc@{\hskip 0.5cm}c}
    \toprule
    {} & \multicolumn{22}{c}{\real{}} && PCN\\
    \cmidrule(r{0.5cm}){2-24}  \cmidrule(r{0.1cm}){25-25} & Ch0 & Ch1 & Ch2 & Ch3 & Du0 & Du1 & Du2 & Du3 & Du4 & Hu0 & Hu1 & Hu2 & Hu3 & Hu4 & Hu5 & Hu6 & Hu7 & Sn0 & Sn1 & Sn2 & Sn3 & Mean (L1) &Mean (L2)& Mean (L1)\\
    \midrule
    ODGNet & 96  & 91  & 90  & 100  & 170  & 172  & 208  & 154  & 191  & 89  & 114  & 106  & 120  & 116  & 120  & 134  & 119  & 83  & 67  & 79  & 80  & 119 & 111 &6\\
    PoinTr & 110 & 113 & 101 & 104 & 99  & 113  & 98   & 92   & 105  & 191 & 146  & 165  & 106  & 132  & 194  & 94   & 121  & 117  & 151  & 144  & 151  & 114 & 58 & 8\\
    AdaPoinTr & 65  & 71  & 71  & 68   & 57   & 54   & 52   & 98   & 37   & 158  & 132  & 114  & 75   & 79   & 134  & 67   & 86   & 65   & 47   & 61   & 53   & 69 & 26 & 7\\
    FoldingNet & 158 & 163 & 181 & 246  & 159  & 143  & 203  & 259  & 161  & 181  & 205  & 154  & 221  & 183  & 235  & 219  & 211  & 142  & 107  & 102  & 74   & 167 & 127& 14\\
    PCN & 145 & 136 & 132 & 123  & 158  & 146  & 167  & 181  & 118  & 170  & 129  & 150  & 139  & 126  & 196  & 118  & 151  & 148  & 106  & 125  & 113  & 143 & 92& 10\\
    TopNet & 563 & 385 & 200 & 211  & 88   & 77   & 90   & 91   & 45   & 590  & 110  & 328  & 94   & 79   & 441  & 93   & 93   & 455  & 75   & 320  & 129  & 341 & 1103 & 12\\
    SnowflakeNet & 58  & -   & 49  & -    & 54   & 73   & 57   & -    & -    & -    & -    & -    & 68   & -    & -    & -    & -    & -    & 57   & -    & 82   & \textbf{60} & 72 & 7\\
    GRNet & 79  & 80  & 79  & 104  & 83   & 63   & 66   & 111  & 64   & 140  & 133  & 131  & 92   & 74   & 155  & 74   & 82   & 81   & 71   & 94   & 73   & 84 & 27 & 9\\
    AnchorFormer & 64  & 68  & 75  & 115  & 58   & 51   & 62   & 84   & 36   & 146  & 131  & 108  & 79   & 85   & 151  & 70   & 92   & 78   & 79   & 75   & 91   & 72 & \textbf{28} &7\\
    \bottomrule
    \end{tabular}
    }
    \vspace{-0.3cm}\caption{Performance of baselines on \real{} dataset. We also show the corresponding average values for the PCN dataset. Despite remarkable performance on current existing datasets, the baselines fail on \real{}. We use Chamfer distance ($\times 10^{-3}$).}
    \label{tab:baseline_comparison_21}
\end{table*}

\begin{figure}
    \centering
    \includegraphics[width=0.5\textwidth, trim={0cm, 4.5cm, 13cm, 5.5cm}, clip]{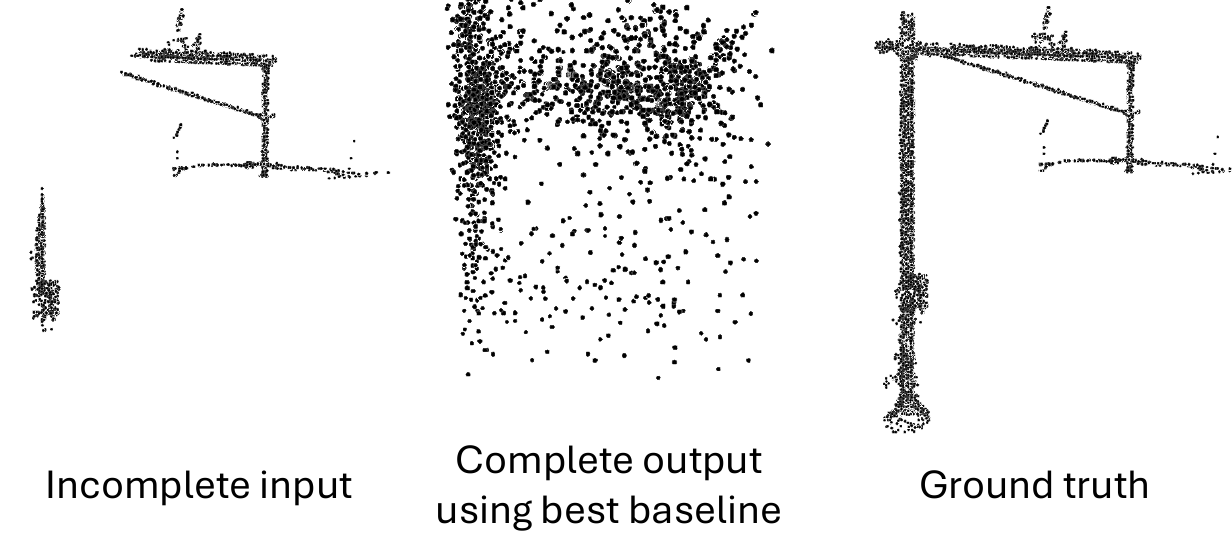}
\caption{Sub-optimal completion results on a \real{} PC using the best-performing completion baseline SnowflakeNet (Table \ref{tab:baseline_comparison_21}).}

\label{fig:base_recon_demons}
\end{figure}

In this subsection, we study and compare \real{}'s performance against existing datasets on three neural methods: (a) Completion, (b) Reconstruction, and (c) Generation. PC completion is a fundamental task in PC perception-based challenges, and it will serve as our primary focus. We benchmark several completion methods on \real{}: (a) ODGNet \cite{ODGNet}, (b) PointTr \cite{pointtr}, (c) AdaPoinTr \cite{pointtr}, (d) FoldingNet \cite{foldingNet}, (e) PCN \cite{yuan2018pcn}, (f) TopNet \cite{topnet}, (g) SnowflakeNet \cite{xiang2021snowflakenet}, (h) GRNet \cite{xie2020grnet}, and (i) AnchorFormer \cite {chen2023anchorformer}. These results are demonstrated in Table \ref{tab:baseline_comparison_21} and Figure \ref{fig:base_recon_demons}.

In Table \ref{tab:benchmark_reconstruction_gen} (Right), we further benchmark \real{} on the task of reconstruction using standard LIDAR backbones: (a) DGCNN \cite{wang2018dynamic}, (b) FoldingNet\cite{yang2018foldingnet}, and (c) the best performing completion method (Table \ref{tab:baseline_comparison_21}), i.e. SnowflakeNet. 

Finally, we benchmark \real{} on a shape generation strong diffusion-based model LION \cite{vahdat2022lion} in Table \ref{tab:benchmark_reconstruction_gen} (Left). 

 
\real{} results on the completion task (Table \ref{tab:baseline_comparison_21} and Figure \ref{fig:base_recon_demons}), and on generation and reconstruction tasks (Table \ref{tab:benchmark_reconstruction_gen}) demonstrate the complexity of \real{} and the performance gap between it and existing object PC datasets.




We discuss the reasons behind the unsatisfactory performance of the baselines as follows: (a) We investigate existing PC datasets and \real{} using tools from algebraic topology and TDA in Section \ref{sec:compar_analysis}. Our discoveries, though striking, are not surprising. It suggests that \real{} consists of non-trivial 0- and 1-dimensional Homology-based topological features that correspond to connected components, and cycles - indicated by the non-diagonal nature of the persistence diagram in Figure \ref{fig:nonuni_topo}. These add to the complexity of \real{} and provide strong evidence of the existence of higher dimensional topological features. We demonstrate that adding these Homology priors as constraints to an existing model can improve completion to some degree (Sec. \ref{sec:odg_PH_model} and Table \ref{tab:odg-topo}). (b) PCs in \real{} are collected from four different sensors. This is a \textit{realistic setting} where different incomplete PCs have been acquired with sensors with different intrinsic characteristics. Hence, different PC samples in such a case may follow different distributions, which is not true for existing datasets. (c) Section \ref{sec:compar_analysis} shows that \real{} has several characteristics that make them challenging to work with. The point cloud acquisition in the real-world is affected by multiple ungovernable parameters. These factors introduce noise, non-uniform and inconsistent patterns - a characteristic that is absent in current object PC datasets.

\section{Methodology}
\subsection{Persistent Homology Regularized Completion}
\label{sec:odg_PH_model}
Section \ref{sec:compar_analysis} establishes that \real{} consists of rich topological features unlike existing datasets. 
We now focus on validating the use of \textit{0-dim} $\mathcal{PH}$ priors on top of an existing benchmarked method, to improve completion performance. We choose a PC completion model that can be easily integrated with $\mathcal{PH}$. There exist certain criteria that allow an existing model to be easily integrated with $\mathcal{PH}$.
A challenge with $\mathcal{PH}$ is that it is compute-intensive. The creation of the Vietoris-Rips complex on a PC and persistence computation is time and compute intensive - number of simplices in the complex increase \textit{exponentially} with increase in the PC size. Models that transform the input PC into seed PCs (at an intermediate model layer) at lower resolutions, generating \textit{multiple} and \textit{sparse} PC seeds, can allow \textit{diverse} and compute-efficient calculation of $\mathcal{PH}$ priors. Such models are suited for integration of $\mathcal{PH}$ priors. 
We find ODGNet \cite{ODGNet} to be a feasible model with multiple sparse seed point clouds at the decoder, which can be easily integrated with topological priors using moderate compute. The availability of these sparse seeds as learnable parameters at intermediate layers helps in training ODGNet using $\mathcal{PH}$-based topological priors.



\subsubsection{Architecture}

\begin{figure}
    \centering
    \includegraphics[width=0.5\textwidth,trim={0cm, 61cm, 28cm, 0cm}, clip]{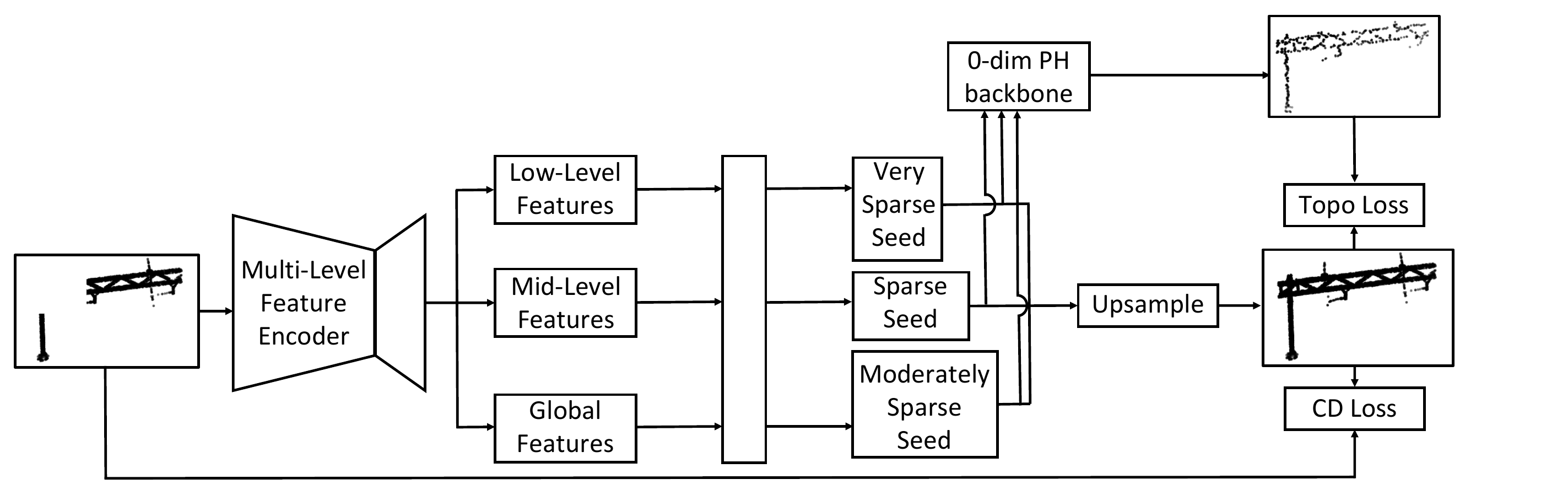}
    \caption{\topodg{}. We calculate \textit{0-dim} $\mathcal{PH}$ based topological priors over sparse seeds and integrate it into the loss function. It enables completion along a topologically consistent skeleton.}
    \label{fig:arch_dia}
\end{figure}

We describe our methodology for the integration of ODGNet with topological priors - \topodg{}. We extract the multi-level seeds from the mid-level, low-level, and global features generated at the decoder. Seed point clouds consisting of points in the range of $256$ to $1024$ can be extracted at the decoder (Figure \ref{fig:arch_dia}). The sparse nature of these seeds makes them easy to work with, and also allows efficient computation of topological features. 

\subsubsection{Topological Loss}
\label{sec:topo_loss}
\begin{figure}[h]
\setlength{\tabcolsep}{1.5pt}
\setlength{\belowcaptionskip}{-3pt}
\begin{tabular}{c c c} 
\includegraphics[width=0.32\linewidth,height=2.5cm]{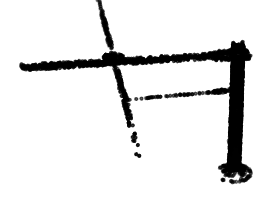} & \includegraphics[width=0.32\linewidth,height=2.5cm]{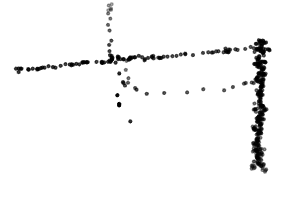}
\includegraphics[width=0.32\linewidth,height=2.5cm]{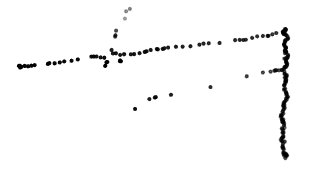}
\end{tabular}

\begin{tabular}{c c c} 
\includegraphics[width=0.32\linewidth,height=2.35cm]{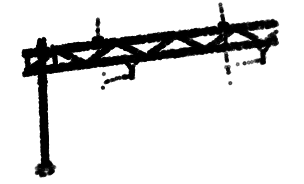} & \includegraphics[width=0.32\linewidth,height=2.35cm]{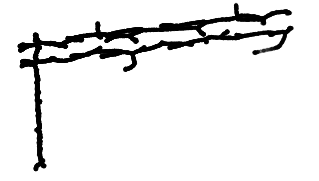}
\includegraphics[width=0.32\linewidth,height=2.35cm]{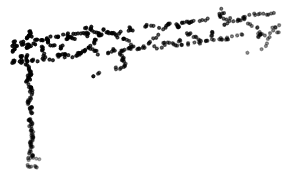}
\end{tabular}

\vspace{-0.3cm}
\caption{\textbf{Left}: A complete PC \textbf{Middle}: Visualization of the $0$-dim $\mathcal{PH}$ based skeleton on a sparse seed during initial training. \textbf{Right}: $0$-dim $\mathcal{PH}$ skeleton using sparse seeds as the training progresses. These priors assist the model in generating topologically consistent complete PCs.
}
\label{fig:skeleton}
\end{figure}


Given an incomplete PC, our goal is to augment it with new points that follow the global topology of the ground truth complete PC. A part of this work is handled by the dictionary module of ODGNet \cite{ODGNet}. It ensures that the seed PCs are able to capture the artifacts of the complete PC and look similar to it. We extract the sparse multi-level PC seeds at the decoder (Figure \ref{fig:arch_dia}). We now apply topological regularization priors on the seed PCs. To obtain this global topology-based backbone we use \textit{0-dim} $\mathcal{PH}$ priors on the seed PCs. \textit{0-dim} $\mathcal{PH}$ ensures the extraction of a complete PC skeleton (Fig. \ref{fig:skeleton}) that can serve as a prior which the network can follow, for complete PC generation. The underlying idea is to ensure shape consistency of the complete output along the skeleton, by generating points along the skeleton. 
We explain the process that extracts \textit{0-dim} $\mathcal{PH}$ priors from the seeds. The seed PC is converted into a simplicial complex. The initial complex is just the original PCs with an $\alpha$-radius ball (here $\alpha$=0). Increasing $\alpha$ adds edges, faces to the simplicial complex, thereby leading to the evolution and death of \textit{k-dim} topological features. Each \textit{k-dim} topological feature is characterized by a \textit{(birth, death)} pair. 

Our goal is to use \textit{0-dim} $\mathcal{PH}$ based topological features. These focus on the global topology of the PC, providing a global skeleton that outlines the complete PC (Figure \ref{fig:skeleton}).

Each \textit{0-dim} homology feature generated using filtration consists of a separate set of \textit{(birth, death)} pairs denoted as \textit{b, d}. The persistence of each pair is \textit{(b-d)} (Section \ref{sec:background}). 

We now explain our topological loss function. For \textit{0-dim} $\mathcal{PH}$, we minimize the sum of persistence for all extracted \textit{0-dim} persistence pairs. This ensures that at the end of the filtration, there is one connected component left (we minimize the persistence of all the pairs \textit{and} there always exists at least one default component in the persistence diagram). This persistence diagram outlines the skeleton of the complete PC, which when used as a regularizer to a model, guides point generation along the skeleton (Figure \ref{fig:skeleton}). This is a vital observation - the network is able to observe \textit{sparse} precise backbones of the complete PC during training which allow it to generate points more closer to the backbone. This observation is of immense interest and we utilize it in the next section to generate accurate complete PCs. 
Given $(b_i, d_i)$ refering to \textit{0-dim} $\mathcal{PH}$ based $(birth, death)$ pairs of a given seed PC, the topology loss is defined as follows: 
\begin{equation}
\label{eq:topo_loss}
        Topo \; Loss = \sum_{i=0}^n  1\{i>k\}(b_i-d_i)
       \\
       = \sum_{i=k+1}^n (b_i-d_i)
\end{equation}

We explain the significance of $k$ now. The input partial PC may not necessarily have a single component i.e. it may be split into more than one component (Fig. \ref{fig:arch_dia}). In such cases, $k\geq$ 2, which allows $\mathcal{PH}$ to generate multiple skeletons, each of which can attend to partial components in the input.

To establish that \textit{0-dim} $\mathcal{PH}$ priors demonstrate a similar effect on other real world datasets, we test these priors on an extensive scene-level dataset (KITTI) for scene completion and reconstruction (details in the supplementary material).

\subsection{BOSH-\underline{B}ackbone \underline{O}utline \underline{S}ampler for P\underline{H}}

In Section \ref{sec:odg_PH_model}, we studied the effect of \textit{0-dim} $\mathcal{PH}$ induced backbones as priors. A major shortcoming of \ref{sec:topo_loss} is the computational complexity of $\mathcal{PH}$. A single backbone requires generating the Vietoris-Rips complex of the PC (Figure \ref{fig:filtration_main}). This is time and compute intensive as the number of simplices increase exponentially with the PC size. Further, the benefit of \ref{sec:topo_loss} is minor, as will be seen in Table \ref{tab:odg-topo} in Section \ref{sec:experiments} later.

\textit{0-dim} backbone priors are proxies to ground truth shapes. We hypothesize that these can be directly sampled from the complete shapes' surfaces at multiple sparsity levels. We circumvent the costly $\mathcal{PH}$ computations by introducing a simple and novel \textit{Homology Sampler} (\textit{BOSH}). \textit{BOSH} directly samples numerous backbones from the surface of the ground truth PC (Figure \ref{fig:arch_dia_2}). These act as proxies to the global \textit{0-dim} $\mathcal{PH}$ backbone. These sampled backbones also serve as seeds to a model, which grow into a complete shape PC as the training proceeds. Introduction of these seed backbones serves a dual purpose: (a) It enables the model to have a fine-grained overview of the complete shape at various resolutions. Given that sparse backbones are more difficult to complete, compared to the original shape, it forces the model to generate explicit attention to the precise shape details of the backbone for completion. (b) This strategy allows circumventing costly Vietoris-Rips complex computations. 

 Let $\{ c_i, p_i \}_{i=1}^n$ be the set of complete and incomplete PC pairs and $BOSH$ be the  \textit{0-dim} Homology Sampler. \textit{M} (here Chamfer Distance) refers to the similarity metric used, \textit{Net} refers to the model and \textit{k} refers to the number of Homology backbones sampled. The new total loss function is given by:

\begin{figure}[t]
    \centering
    \includegraphics[width=0.5\textwidth,trim={0cm, 60cm, 30cm, 0cm}, clip]{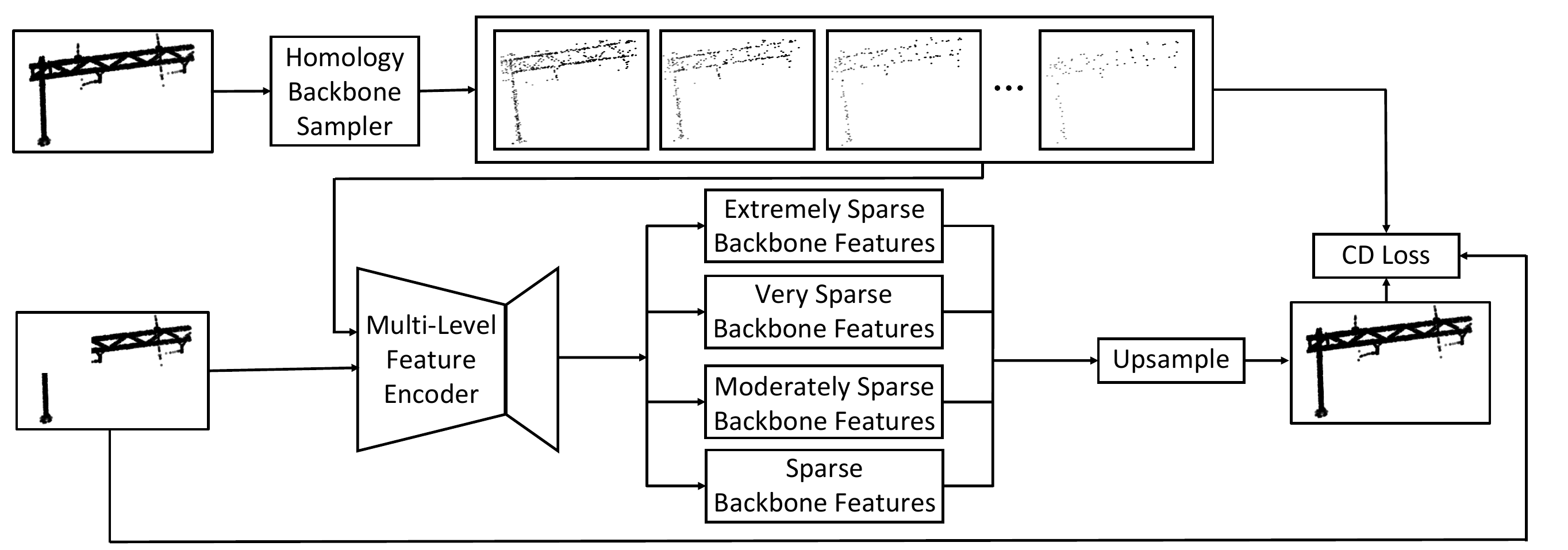}
    \caption{\bosh{}. Our compute-efficient Homology Sampler samples proxy $\mathcal{PH}$ backbones from the surface of the complete scan. These guide the completion process from the start of training.}
    \label{fig:arch_dia_2}
\end{figure}

\vspace{-0.5cm} 
\begin{equation}
\small
\begin{split}
    \sum_{i=1}^{n}\sum_{j=1}^{k} M(Net(BOSH(c_i,j)),c_i ) + \sum_{i=1}^{n} M(Net(p_i),c_i ) 
    \end{split}
\end{equation}

\section{Experiments}

\label{sec:experiments}
\begin{table}[h!]
\small
\centering
\begin{tabular}{lcc}
\textbf{Method} & \textbf{CD-L1} & \textbf{CD-L2} \\ 
\hline
ODGNet & 119 & 111 \\ 
    \topodg{} & 103& 80 \\ 
    SnowFlakeNet  & \textbf{60} & 72 \\ 
\bosh{} & 69 & \textbf{5.4} \\ 
\hline
\end{tabular}
\vspace{-0.3cm}
\caption{Performance comparison of  \topodg{} and \bosh{} against the baseline models.}
\label{tab:odg-topo}
\end{table}

\begin{figure}
    \centering
    \includegraphics[width=0.47\textwidth, trim={0cm, 6.2cm, 12cm, 0cm}, clip]{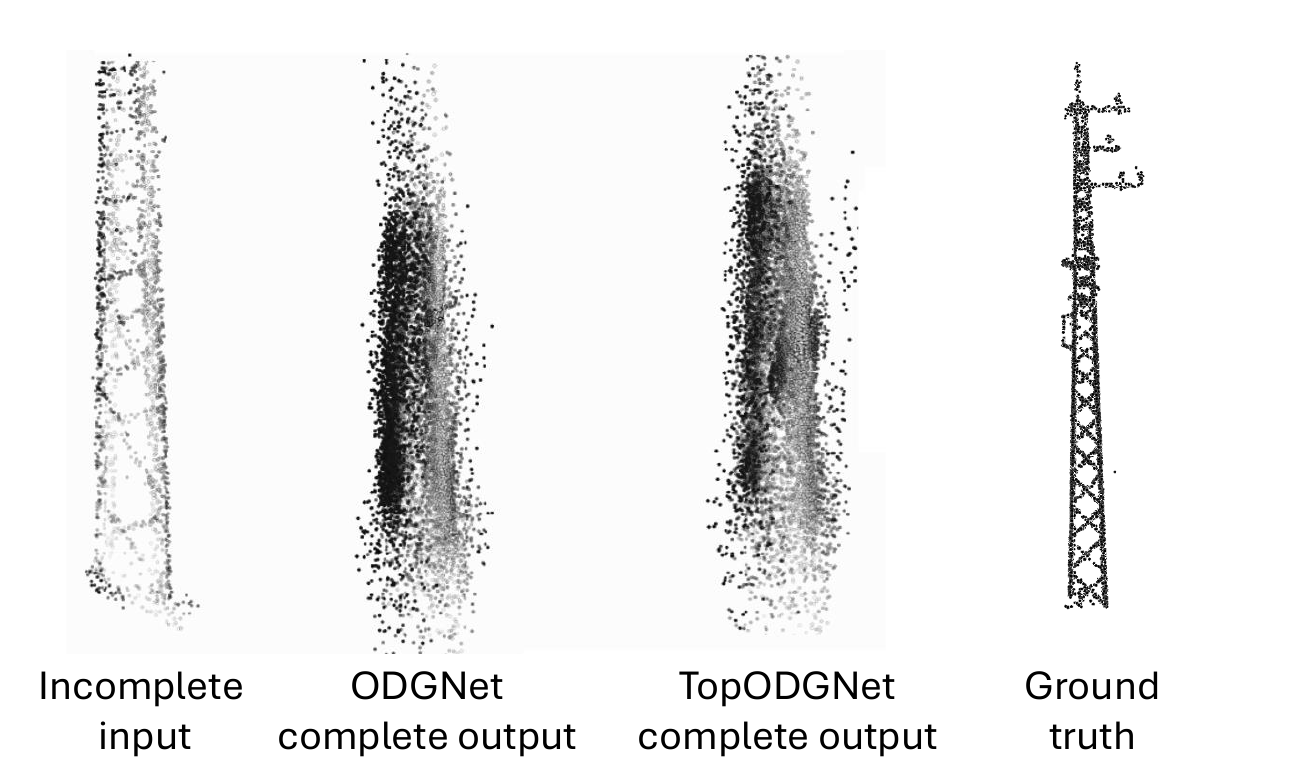}
    \caption{We find that  \topodg{} is able to follow the global topology of the ground truth PC better than ODGNet, as indicated by the tapering upper half of the output and overall consistency. \topodg{} shows visible benefits of introducing topological priors.}    \label{fig:completion}
\end{figure}

\begin{figure}
    \centering
    \includegraphics[width=0.5\textwidth, trim={0cm, 4.2cm, 16cm, 1cm}, clip]{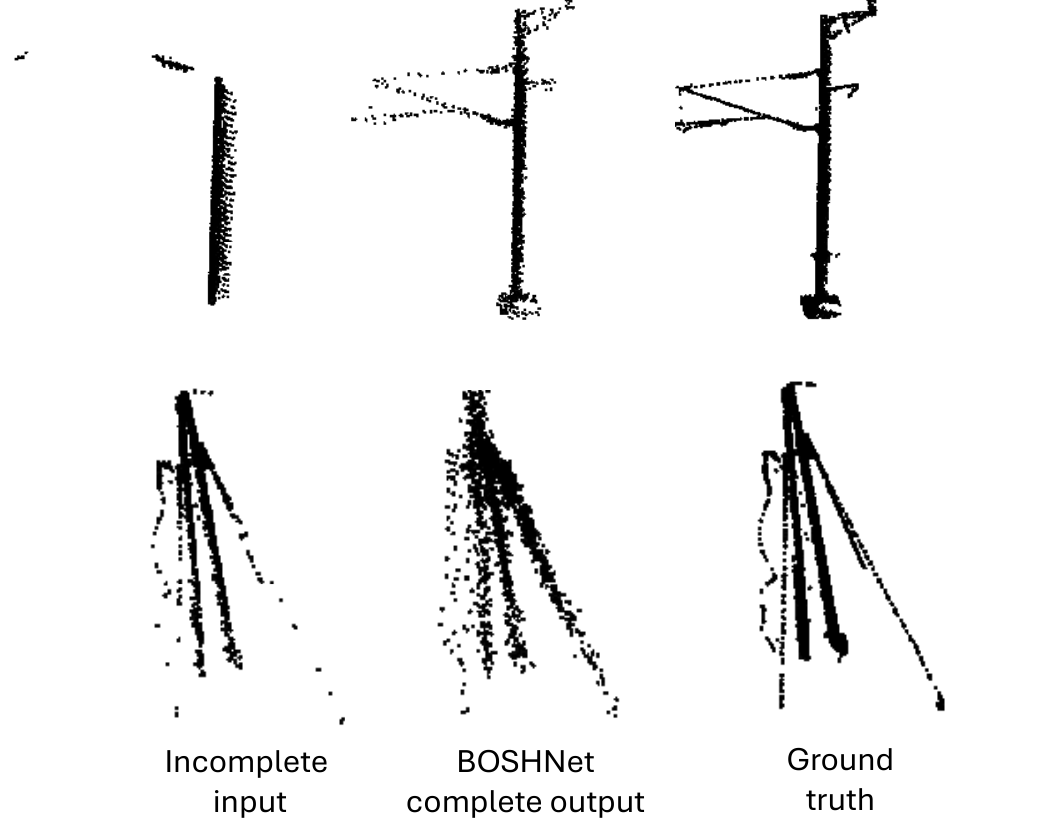}
\caption{\bosh{}, on account of multiple \textit{0-dim} $\mathcal{PH}$ priors, is able to reconstruct the incomplete PC reasonably.
}
\label{fig:homology-sampler}
\end{figure}



\vspace{-0.4cm}
We demonstrate the benefit of topological priors on the ODGNet backbone. Figure \ref{fig:completion} shows the relative topological consistency of \topodg{} \textit{w.r.t.} the baseline. This results in a slight reduction of Chamfer distance as shown in Table \ref{tab:odg-topo}. While both the outputs are still noisy, \topodg{} maintains a better topological consistency \textit{w.r.t.} the ground truth.

We demonstrate the effectiveness of our Homology Sampler Completion Network, \bosh{} in Figure \ref{fig:homology-sampler} and Table \ref{tab:odg-topo}. The Homology Sampler outperforms all baselines by a significant margin for CD-L2. It performs comparable to the best baseline for CD-L1. A vital benefit of our approach is that the network has access to the multiple \textit{0-dim} $\mathcal{PH}$ prior backbones from the start of the training, unlike the case with \topodg{}, which depends on the learned sparse seeds to generate the backbone. These seeds are noisy during the start of the training and sharpen only during the later stages of the training.
\section{Conclusion}

We discover that real-world PCs in uncontrolled settings have non-uniform density, noise, and rich and versatile topological features (extracted using tools from Algebraic Topology). These are non-existent in existing datasets. Existing methods for PC completion do not account for these and fail miserably for real-world PCs.  We introduce a topologically-rich real-world PC completion dataset, \real{} with 21 categories across $\sim$ 40,000 pairs. We benchmark several state-of-the-art baselines on \real{} and demonstrate the need to rethink PC completion for the real-world. We demonstrate that utilizing $\mathcal{PH}$-based topological features as priors for real-world PCs can help in generating topologically accurate complete PCs.

{\small
\bibliographystyle{abbrv}
\bibliography{mainTemplatePDF}
}

\clearpage
\setcounter{page}{1}
\maketitlesupplementary

\section{Persistent Homology}

\begin{figure*}
    \centering
\includegraphics[width=\textwidth, trim={0cm, 65cm, 35cm, 0cm}, clip]{figures/filt.pdf}
\caption{(a) to (c) Progression of filtration on a point cloud over different spatial resolutions as the distance threshold increases \cite{moor2020topological}. (d) Birth and death of $k$-dim topological features documented in the form of a persistence diagram, i.e., $(b_i,d_i)$ pairs, so that each point corresponds to a homology which is born at $b_i$ and dies at $d_i$.}
\label{fig:filtration}
\end{figure*}

In this section, we delve into the concept of Persistent Homology ($\mathcal{PH}$), an important tool in topological data analysis that systematically uncovers and quantifies the topological features of datasets. Persistent Homology combines ideas from algebraic topology, geometry, and computational mathematics to identify meaningful structures, such as connected components, loops, and voids, that persist across multiple scales. It is particularly valuable when applied to data in the form of point clouds or pixelated images, where traditional methods might struggle to capture structural and relational information.

At the core of this process lies the representation of a topological space as a cell complex—a combinatorial structure that encodes the relationships between points in the space. These cell complexes are constructed using simplices, which are the building blocks of higher-dimensional shapes. A 0-dimensional simplex is a point, a 1-dimensional simplex is an edge, a 2-dimensional simplex is a triangle, and so on, with higher-dimensional simplices being generalizations of these structures. Simplices are combined to form simplicial complexes, which generalize graphs to higher dimensions. Figure \ref{fig:filtration} provides a visual depiction of simplices of various dimensions and how they combine to form simplicial complexes. These structures serve as the foundation for applying homological methods.

Homology, in its simplest sense, is a branch of mathematics that provides a systematic way to analyze and classify the global properties of a topological space by examining its local features. Specifically, homology assigns algebraic invariants to a topological space, allowing us to identify and quantify $k$-dimensional holes, where $k$ corresponds to the dimension of the feature being analyzed. For example:
a 0-dimensional hole corresponds to a connected component of the space,
 a 1-dimensional hole corresponds to a loop or cycle, a 2-dimensional hole corresponds to a void or cavity enclosed by a surface.

These holes, generalized across all dimensions, encapsulate the structural essence of a space. Figure \ref{fig:filtration} illustrates how homology captures these features, translating the geometry of a space into meaningful topological features \cite{edelsbrunner2008computational}.

Persistent Homology ($\mathcal{PH}$) extends classical homology by tracking how these $k$-dimensional features change as the dataset is viewed at different scales. This is accomplished by constructing a filtration, which is a sequence of nested simplicial complexes:
\[
\phi \subseteq \mathcal{C}_1 \subseteq \mathcal{C}_2 \subseteq \mathcal{C}_3 \dots \mathcal{C}_i \dots \mathcal{C}_n = \mathcal{C},
\]
where $\phi$ is the empty complex, and $\mathcal{C}$ is the final simplicial complex encompassing the entire dataset. Each step in this sequence corresponds to a specific scale or resolution. Persistent Homology identifies when topological features, such as connected components, cycles, or voids, are born and when they die as the filtration progresses. Features that persist across a wide range of scales are considered significant and indicative of meaningful structures within the data. Conversely, features that appear and disappear quickly are often interpreted as noise \cite{edelsbrunner2008computational}.

Filtration is the step-by-step process of building simplicial complexes by progressively adding simplices to the structure. The manner in which this process is defined depends on the type of dataset being analyzed. For example, in the case of point clouds, a filtration is often constructed using distances between points. At each stage of the filtration, new simplices are added based on a chosen threshold parameter, often denoted by $\alpha$. The filtration progresses monotonically, meaning that each simplicial complex in the sequence contains all simplices from the previous step, along with any new simplices added at that stage.

For point clouds, a common type of filtration is the Vietoris–Rips filtration, which is defined based on pairwise distances between points. For a given value of $\alpha$, an edge is added between two points if the distance between them is less than or equal to $2\alpha$. Higher-dimensional simplices, such as triangles and tetrahedra, are introduced when a set of points becomes fully connected. The progression of this filtration is depicted in Table \ref{fig:filtration}, which illustrates how simplicial complexes evolve as $\alpha$ increases.

As the filtration progresses, topological features are born and die. These events are recorded as pairs $(b, d)$, where $b$ is the scale at which the feature first appears (birth), and $d$ is the scale at which the feature disappears (death). For example:
- The addition of an edge may create a new 1-dimensional cycle, marking the birth of a feature.
- Conversely, the addition of another edge may fill in that cycle, causing its death.

Consider the example of four points forming a rectangle. Initially, a 1-dimensional hole (cycle) is created when the rectangle is formed. When the diagonal edge is added, the rectangle is divided into two triangles (2-simplices), resulting in the destruction of the cycle. These birth-death pairs can be visualized using barcodes, where the length of each bar represents the persistence of a feature. Long bars correspond to significant features, while short bars typically represent noise \cite{edelsbrunner2008computational}.

When dealing with images, the process of constructing a filtration differs from that of point clouds. Instead of using pairwise distances, the pixel intensities of the image are used. The final simplicial complex $\mathcal{C}$ corresponds to a triangulation of the image grid, with vertices representing pixels. Sub-level set filtrations are used, where the filtration function is defined as:
\[
f((v_0,v_1...v_n)) = \max\limits_{i=0,1,2,3...n }f(v_i),
\]
which assigns each simplex a value equal to the maximum intensity of its vertices. Filtration begins with the minimum intensity value and gradually includes pixels with intensities less than or equal to $\alpha$. As $\alpha$ increases, new simplices are added, and the filtration progresses. This allows the topological features of the image to be analyzed at multiple intensity levels.

     
     

Persistent Homology is a powerful framework that extracts meaningful structural information from complex datasets. By studying the persistence of topological features across scales, it provides insights into the underlying geometry and topology of the data. Its versatility makes it applicable to a wide range of domains, including shape analysis, image processing, and network analysis.

\section{Evaluation Metrics}
We evaluate the difference between model reconstructed complete PC and ground truth complete PC using the following metrics.
\begin{itemize}
    
    \item Chamfer Distance (CD): It tries to capture the average mismatch between points in two given point clouds $X, Y \in \mathbb{R}^3$ and is given by:
    \[ d_{CD-L2}(X, Y) = \sum_{x \in X}\min_{y \in Y}||x-y||_2^2 + \sum_{y \in Y}\min_{x \in X}||x-y||_2^2\]

    \[ d_{CD-L1}(X, Y) = \sum_{x \in X}\min_{y \in Y}|x-y| + \sum_{y \in Y}\min_{x \in X}|x-y|\]

    \item Hausdorff Distance (HD): If the distance between a point $p$ on a surface $S$ and a surface $S'$ is given by:

    \[
        d(p, S') = \min_{p' \in S'}||p-p'||_2
   \]
then the HD between $S$ and $S'$ is defined as:

    \[
        d_H(S, S') = \max_{p \in S}d(p,S')
   \]

For the evaluation of our surface reconstruction task (Section 4.3 of the main paper), we use an average of $ d_H(S, S')$ and $d_H(S', S)$.

\end{itemize}

\section{Scene-Level Datasets}
We provide a detailed description of the procedure for creation of \real{}.
To create \real{}, as mentioned in Section 4.1 of the main paper, we use four open-source railway datasets: \cite{cserep2022hungarian} by Hungarian State Railways acquired with a Riegl VMX-450 high-density mobile mapping system; \cite{qiu2024whu} by Wuhan University, in which urban railway dataset was captured using Optech’s Lynx Mobile Mapper System, rural railway dataset with MLS system equipped with HiScan-Z LiDAR sensors, and plateau railway dataset with Rail Mobile Measurement System (rMMS) equipped with a 32-line LiDAR sensor; \cite{sncfNuagePoints} by SNCF Réseau, the French state-owned railway company; and a catenary arch dataset \cite{ton2022labelled} given by Strukton Rail, captured using Trimble TX8 Terrestrial Laser Scanner (TLS). A few scenes from these datasets have been visualized in Figure 3 in the main paper.

We use four open-source scene-level railway datasets from different countries. These scene-level PCs are annotated into relevant sections such as vegetation, overhead cables, railway tracks, industrial support and transmission structures, ground, etc. For this work, we extract industrial structures from these scene-level PCs. As a result, we obtain a PC with multiple industrial structures as shown in Figure 4 in the main paper - Input to (A). Referring to the same figure, our methodology is divided into five main parts. In (A) we employ HDBSCAN \cite{campello2013density}, a hierarchical clustering algorithm, to cluster individual industrial structures. HDBSCAN works accurately for noisy and complex data and automatically detects the number of data clusters. In (B), a manual inspection is performed to extract industrial structures which can serve as good ground truth for supervised training. Finally, in (C), (D), and (E), we process these ground truth PCs in three different ways to induce uniform sparsity, non-uniform sparsity, and incompleteness as explained in detail ahead.

For making a partial PC (C), we pick a random viewpoint around a ground truth PC and remove N number of points that are farthest away from this viewpoint. Similarly for sparsifying the same PC non-uniformly (D), we again pick a random viewpoint but this time we assign a probability to each point, which is either proportional or inversely proportional to a point's cubed distance to this viewpoint. Then, we sample N points based on these probability values. This is repeated for different values of N and for different viewpoints. For uniform sparsification (E), we randomly sample N points and then repeat for different values of N. Simultaneous to these steps, we perform manual noise removal. These three steps are conducted for all ground truth PCs to obtain a paired object-level training dataset.

Steps (B) and (C) have manual interventions. We want to highlight that these interventions make \real{} robust and accurate. Most point clouds extracted from industrial scenes are incomplete due to view occlusions present during acquisition. To ensure that all ground truth point clouds in \real{} are complete in all respects, we manually hand-pick the complete shapes (B). To group samples into different classes, manual inspection and labeling is known to be more accurate as compared to automated processes. We intensively study all available shapes and acquire domain knowledge of the dataset before manually categorizing the point clouds. 

\real{} is a real-world object point cloud dataset generated using mostly scene-level parent datasets. Topological properties of \real{} are inherited from the parent datasets. Our proposed method exploits these properties using $\mathcal{PH}$ priors.

A differentiating factor of \real{} is that - unlike existing datasets (which may or may not be simulated) that are captured in extremely controlled environments, \real{} is acquired in uncontrolled real-world multi-sensor industrial settings that have numerous factors of variation, noise and disturbance. 

We also highlight in the main paper that several existing datasets (mentioned in L129-132) though not simulated, are captured in extremely controlled environments - unlike real-world settings (e.g. industrial) that have numerous factors of variation, noise and disturbance. These datasets (including KITTI) do not consist of ground truth complete shapes and hence do not consist of partial-complete paired information.

\section{Non-Neural Methods}

\label{sec:non-neural}

In subsection 4.3 of the main paper, we benchmark our \real{} and ShapeNet against three non-neural network-based tasks: (a) Simplification, (b) Surface reconstruction, and (c) Upsampling. Point cloud simplification reduces point density of a point cloud while preserving its salient features and the overall structure. This enables faster processing, lower storage costs, and efficient analysis for large-scale 3D data. We use Weighted Locally Optimal Projection (WLOP) \cite{huang2009consolidation} in the mentioned subsection, an enhancement of the parameterization-free denoising and simplification method known as Locally Optimal Projection (LOP) \cite{lipman2007parameterization}. While LOP struggles with point clouds that exhibit non-uniform distributions, WLOP addresses this shortcoming by integrating locally adaptive density weights.

\begin{figure*}[ht]
    \centering

    \begin{subfigure}{\textwidth}
        \centering
        \includegraphics[width=0.97\textwidth, trim={0.5cm, 64cm, 37cm, 0cm}, clip]{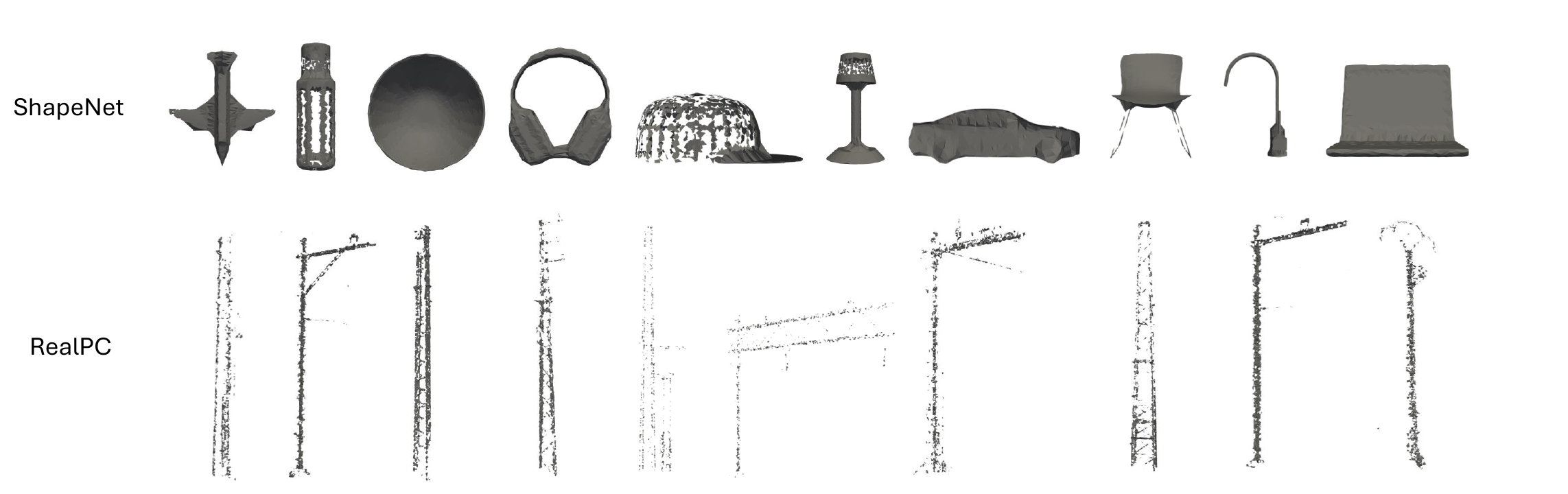}
        \caption{$\alpha = 0.05$}
    \end{subfigure}

    \vspace{1em} 

    \begin{subfigure}{\textwidth}
        \centering
        \includegraphics[width=\textwidth, trim={0cm, 64cm, 35cm, 1cm}, clip]{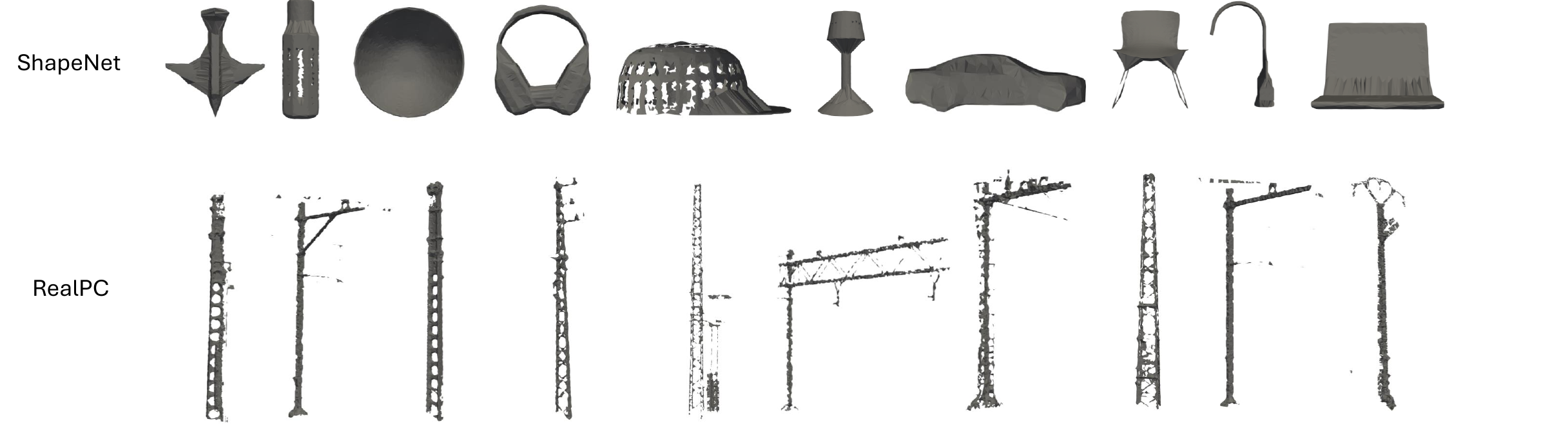}
        \caption{$\alpha = 0.1$}
    \end{subfigure}

    \vspace{1em} 

    \begin{subfigure}{\textwidth}
        \centering
        \includegraphics[width=\textwidth, trim={0cm, 64cm, 35cm, 1cm}, clip]{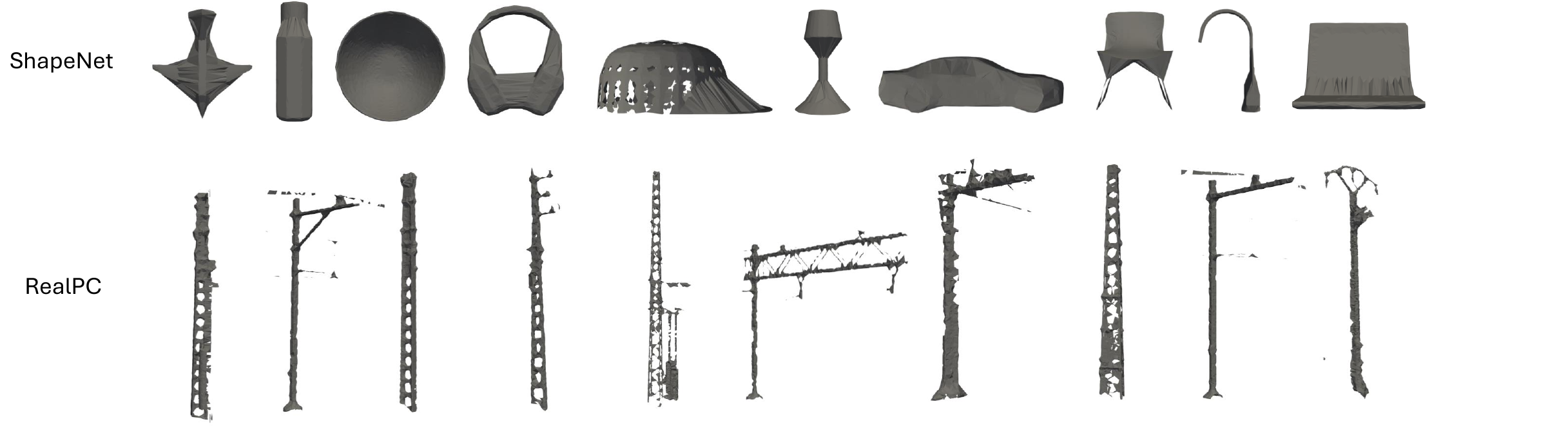}
        \caption{$\alpha = 0.15$}
    \end{subfigure}

    \caption{Surface reconstruction of some instances from different categories of ShapeNet and RealPC using alpha shapes.}
    \label{fig:combined_figures2}
\end{figure*}

Surface reconstruction for point clouds aims to generate a continuous surface from discrete points, enabling the creation of complete 3D models essential for visualization, analysis, and manufacturing. An alpha shape \cite{edelsbrunner1983shape} is basically made from a subcomplex of the Delaunay triangulation of a point cloud, with its refinement level ranging from a rough approximation to a highly detailed representation of the point cloud's surface. For our comparison, we use a range of $\alpha$ values to create a mesh from different point clouds. Mesh visualizations shown in Figure \ref{fig:combined_figures2} support the average high HD values in the case of \real{} as compared to ShapeNet (Table 2 of the main paper). The meshes formed from \real{} point clouds fail to capture the intricate details of the point cloud at any $\alpha$ value.

Point cloud upsampling increases the density of sparse point clouds, enabling finer surface details and improving the accuracy of reconstruction, analysis, and downstream processing tasks. We adopt an upsampling strategy \cite{huang2013edge} which uses an edge-aware method to compute normals away from the surface singularities, and then uses the data obtained to resample towards the said singularities with the help of a bilateral projector.



\section{Topological Loss}

\subsection{Formulation}
The topological loss function assists the completion process by ensuring that the model generates point clouds along the topological prior generated by \textit{0-dim} $\mathcal{PH}$. 

The coarse point cloud seeds at the decoder assist the topological module and are used as input for the topological loss. The output of the topological module consists of \textit{0-dim} topological features (\textit{birth, death}) pairs. The loss function in Equation \ref{eq:topo_loss_supp} below uses these pairs for adjusting the \textit{birth} and death values of the \textit{0-dim} Homology features according to the required skeleton and the available components in the input incomplete point cloud.

\begin{equation}
\label{eq:topo_loss_supp}
        \mathcal{L}_{homo} = \sum_{i=0}^n  1\{i>k\}(b_i-d_i)
       \\
       = \sum_{i=k+1}^n (b_i-d_i)
\end{equation}

The parameter \textit{k} holds significance here. As explained in the main paper, the input partial point cloud may consist of multiple disconnected components as demonstrated in Figure 6 in the main paper. In such cases, the above loss formulation ensures that the skeleton can have three skeleton components that can \textbf{(a)} cater to the three disconnected components and \textbf{(b)} merge together to form the final complete point cloud skeleton. 

Setting the value of \textit{k} requires manual inspection of the partial point clouds. In cases where the majority of point clouds have single components, setting \textit{k=1} works fairly well. 

This formulation assists in the formation of the \textit{0-dim} $\mathcal{PH}$ skeleton easily when the partial point clouds are not continuous (points are in disconnected clusters).

\subsection{PH priors for KITTI scene understanding}
We test these topological priors for scene reconstruction as well as completion on the real-world KITTI Odometry dataset (\href{[https://www.kaggle.com/datasets/prashk1312/kitti-static-dynamic-correpsondence](https://www.kaggle.com/datasets/prashk1312/kitti-static-dynamic-correpsondence)}{data}). We cannot use \topodg{} and its modules (dictionary, seed generation, orthogonality constraint) are optimized for shape completion \cite{ODGNet}. For KITTI scenes, we develop a generative model (inspired by DGCNN) and test it with and without \textit{0-dim} $\mathcal{PH}$ priors. We use DGCNN \cite{wang2019dynamic} because it can easily be used for integrating topological priors as it satisfies the required criterion for integrating $\mathcal{PH}$ priors. It transforms each point of a point cloud using Edge Convolutions into higher dimensions and retains the number of points across each layer. Therefore, sparse seeds point clouds of higher dimensions can easily be extracted from these layers and used for computation of the Vietoris Complex and persistence information. We use Sequence 08 for testing, 03 for validation, and the rest for training DGCNN with the KITTI dataset. The results for the experiments are shown in Table \ref{tab:benchmark_kitti}.

\begin{table}[htbp]
    \centering
    \small 
    \setlength{\tabcolsep}{3pt} 
    \label{tab:pointcloud_tasks}
    \begin{tabular}{c@{\hskip 0.3cm}cc}
    \toprule
    {} & \textbf{Without $\mathcal{PH}$ priors} & \textbf{With $\mathcal{PH}$ priors}  \\
    \cmidrule(){2-3}
     Scene Completion   & 1.18 & \textbf{1.04}  \\
     Scene Reconstruction  & 1.50 & \textbf{1.18} \\
    \bottomrule
    \end{tabular}
\caption{Benchmarking result for scene reconstruction and completion using the KITTI dataset using the Chamfer Distance $\downarrow$ metric. All numbers are ($ \times 10^{-3}$).}
\label{tab:benchmark_kitti}
\end{table}

\begin{figure}[htbp]
\setlength{\tabcolsep}{1.5pt}
\setlength{\belowcaptionskip}{-3pt}
\begin{tabular}{c c c} 
\includegraphics[width=0.32\linewidth,height=3.5cm,trim={3cm, 2cm, 3cm, 1.5cm}, clip]{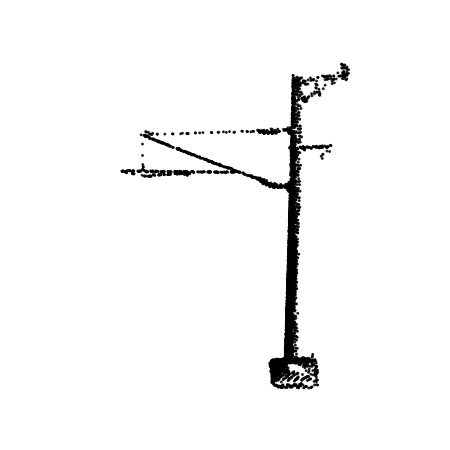} & \includegraphics[width=0.32\linewidth,height=3.5cm,trim={3cm, 2cm, 3cm, 1.5cm}, clip]{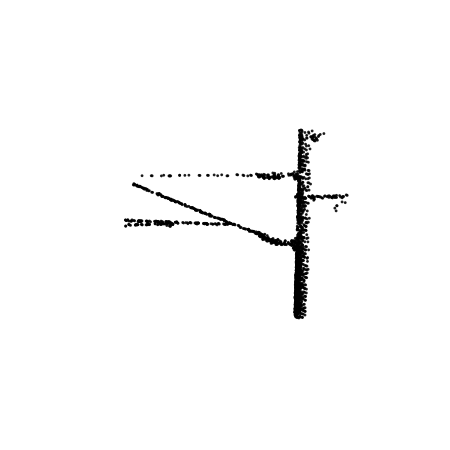}
\includegraphics[width=0.32\linewidth,height=3.5cm,trim={3cm, 2cm, 3cm, 1.5cm}, clip]{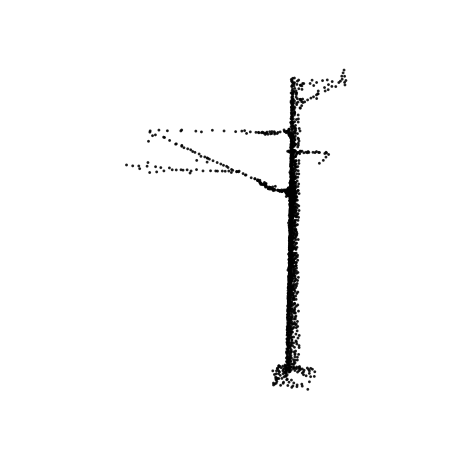}
\end{tabular}

\begin{tabular}{c c c} 
\includegraphics[width=0.32\linewidth,height=3.5cm,trim={3cm, 4cm, 3cm, 2.7cm}, clip]{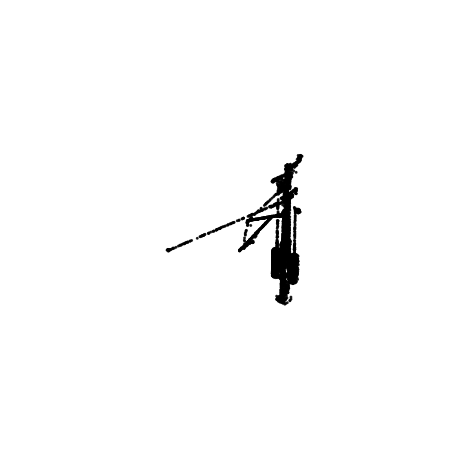} & \includegraphics[width=0.32\linewidth,height=3.5cm,trim={3cm, 4cm, 3cm, 2.7cm}, clip]{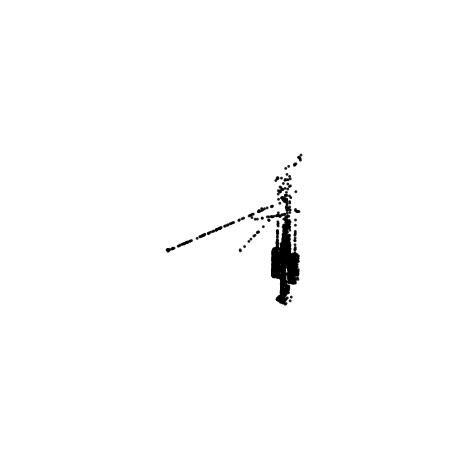}
\includegraphics[width=0.32\linewidth,height=3.5cm,trim={3cm, 4cm, 3cm, 2.7cm}, clip]{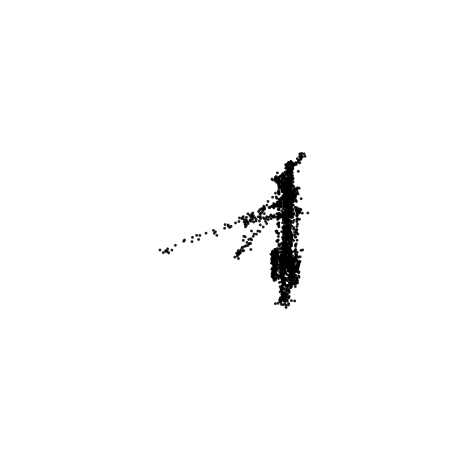}

\end{tabular}
\begin{tabular}{c c c} 
\includegraphics[width=0.32\linewidth,height=3.5cm,trim={3cm, 4cm, 3cm, 2.7cm}, clip]{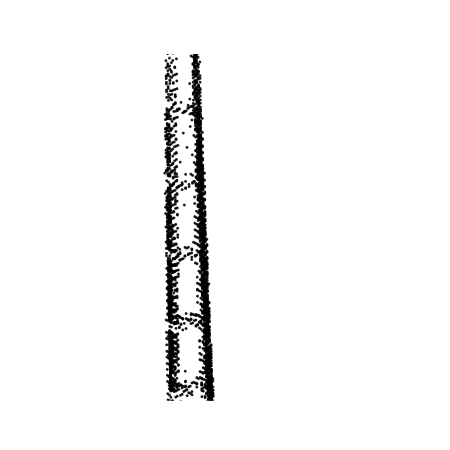} & \includegraphics[width=0.32\linewidth,height=3.5cm,trim={3cm, 4cm, 3cm, 2.7cm}, clip]{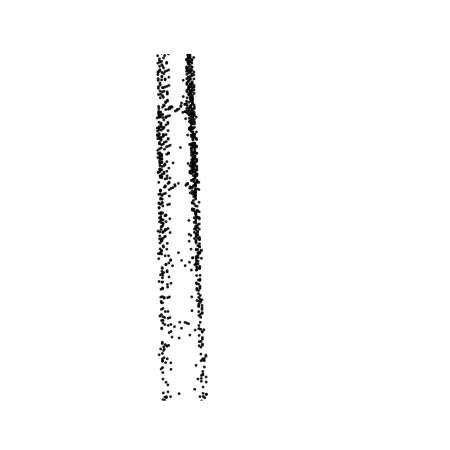}
\includegraphics[width=0.32\linewidth,height=3.5cm,trim={3cm, 4cm, 3cm, 2.7cm}, clip]{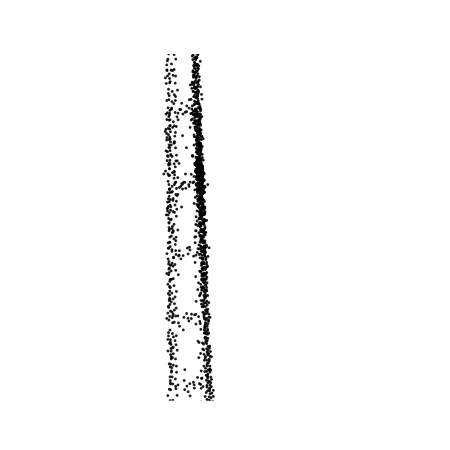}
\end{tabular}

\caption{\textbf{Left}: Complete GT  \textbf{Middle}: Partial Input to Homology Sampler based Model \textbf{Right}: Output.  Our Homology Sampler model, on account of multiple \textit{0-dim} $\mathcal{PH}$ priors is able to accurately reconstruct the incomplete PC.
}
\label{fig:homology-sampler-recon-2}
\end{figure}

\section{Experimental Details}

We provide the experimental details here. Our models are trained using an NVIDIA A100 GPU. 
For training the \topodg{} we use the same parameters and model structure as ODGNet \cite{ODGNet}. We plug in the topology module on the sparse seeds generated at the decoder as demonstrated in Figure 6 in the main paper.

For Homology Sampler based network, we train the network for 1000 epochs with Adam optimizer using an initial learning rate of 5e-4. We use a standard Point Cloud Autoencoder backbone based on PointNet.

\section{Demonstrations}
We visually demonstrate nine point clouds from our dataset in Table \ref{tab:cate_1} and \ref{tab:cate_2}. These demonstrate the non-uniform sparsity and noise that are natural and intrinsic characteristics of real-world datasets. 

We demonstrate the Persistence Diagrams of these nine point clouds in Figure \ref{tab:pd}. For most of the point clouds we observe non-trivial persistence (most points are far from the diagonal). This indicates that real-world point clouds captured in challenging settings exhibit significant zero and one-dimensional topological features which are absent in synthetic point clouds (refer to Table 1 in the main paper). These persistence features are not observed in the persistence diagram of the synthetic datasets as shown in Table \ref{tab:pd_shapenet}. Almost all the 0-dimensional topological features (indicated by dot) are along or very close to the diagonal for the synthetic datasets (Table \ref{tab:pd_shapenet}) as opposed to \real{} (Table \ref{tab:pd}). 

We also show some more examples of \bosh{} completions using \real{} in Figure \ref{fig:homology-sampler-recon-2}. \bosh{} picks up precise topology of the point clouds and generates decent complete point clouds.

\begin{table*}
\centering
\setlength{\tabcolsep}{7pt}
\begin{tabular}{c c c c}

\includegraphics[width=0.26\linewidth, height=0.6\linewidth  ]{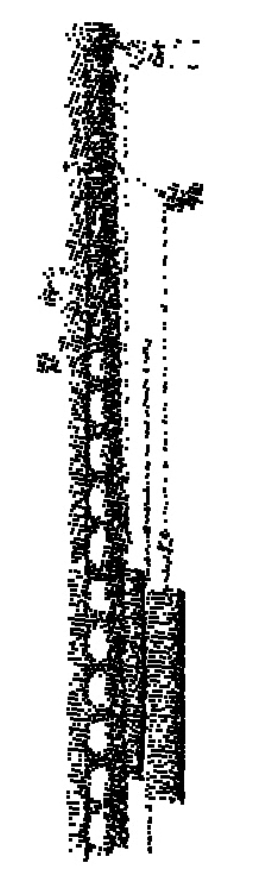}& \includegraphics[width=0.26\linewidth, height=0.6\linewidth ]{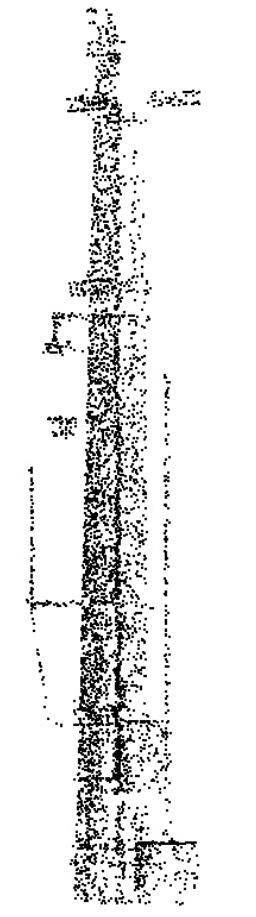}& \includegraphics[width=0.26\linewidth, height=0.6\linewidth ]{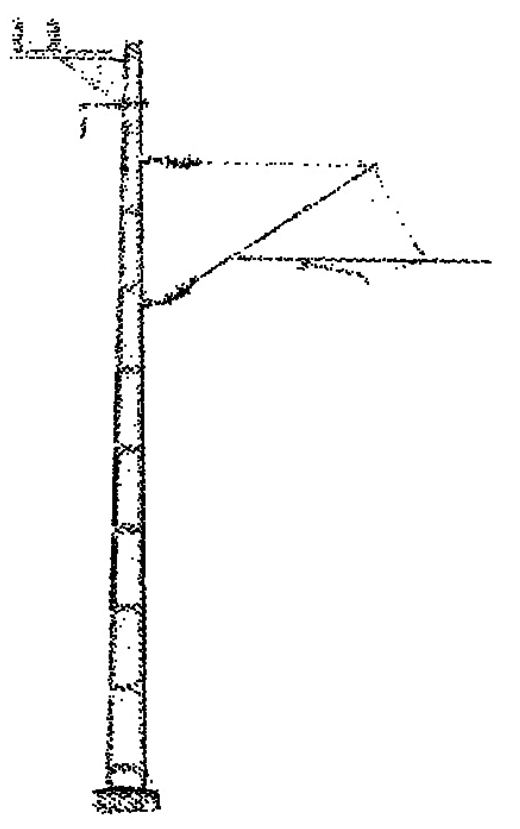} \\ 

\includegraphics[width=0.26\linewidth, height=0.6\linewidth ]{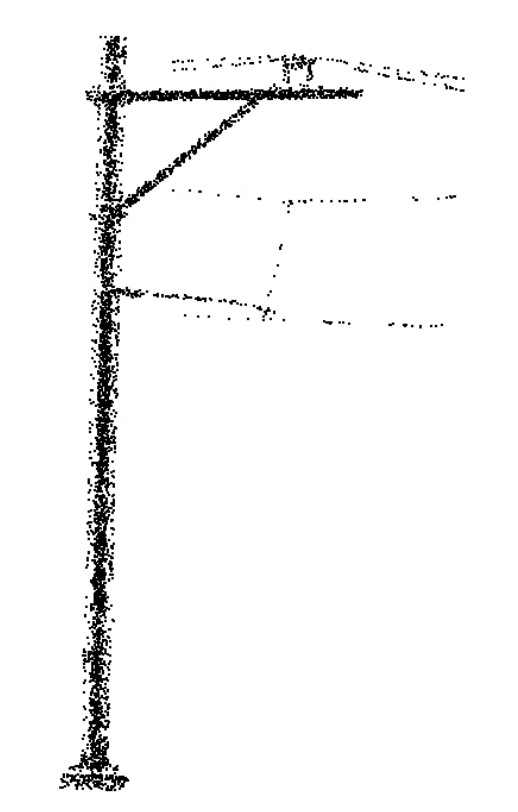}
&
\includegraphics[width=0.26\linewidth, height=0.6\linewidth ]{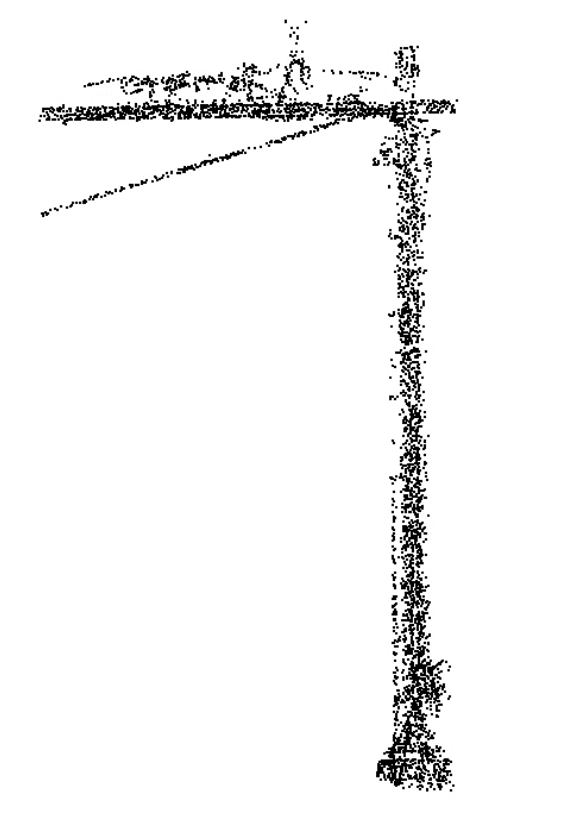}& \includegraphics[width=0.26\linewidth, height=0.6\linewidth  ]{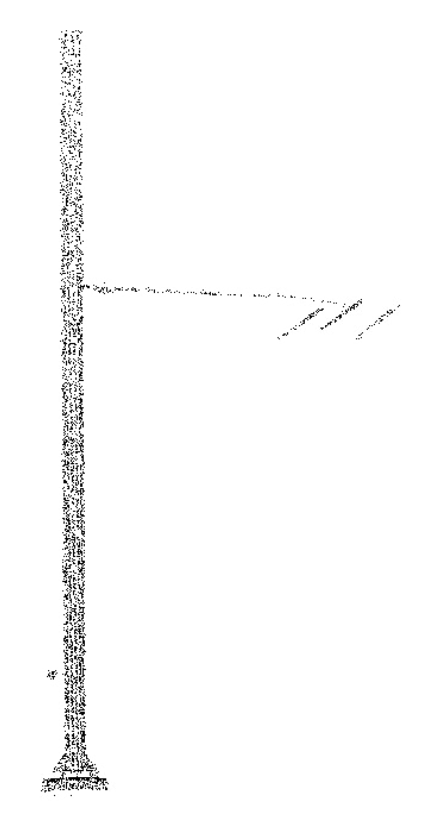} \\ 
 
\end{tabular}
\caption{Visual Demonstration of some examples from our dataset. For a video demonstration please refer to the video in the supplementary.}
\label{tab:cate_1}
\end{table*}
\begin{table*}
\centering
\setlength{\tabcolsep}{7pt}
\begin{tabular}{c c c c}
 \includegraphics[width=0.26\linewidth, height=0.6\linewidth ]{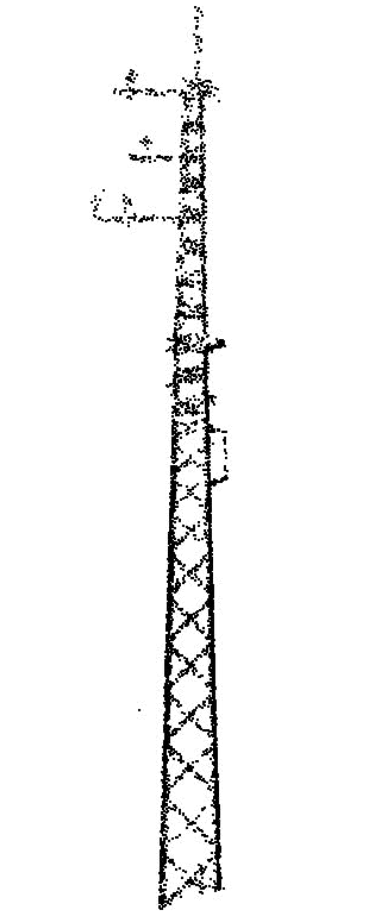}
 &\includegraphics[width=0.26\linewidth, height=0.6\linewidth  ]{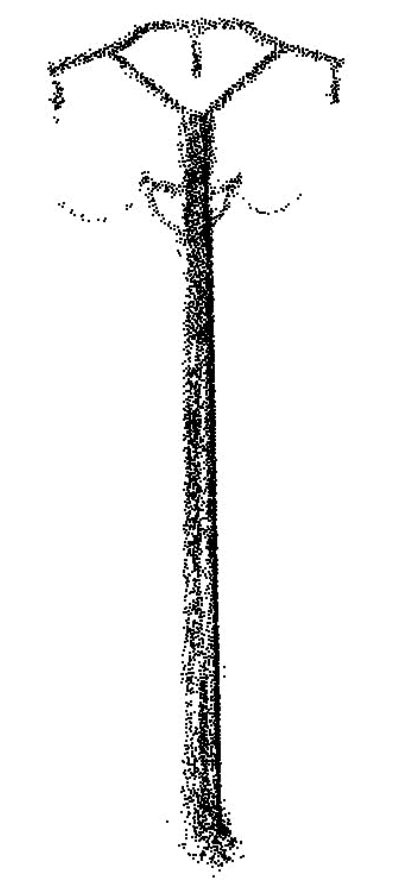}
 & \includegraphics[width=0.26\linewidth, height=0.6\linewidth  ]{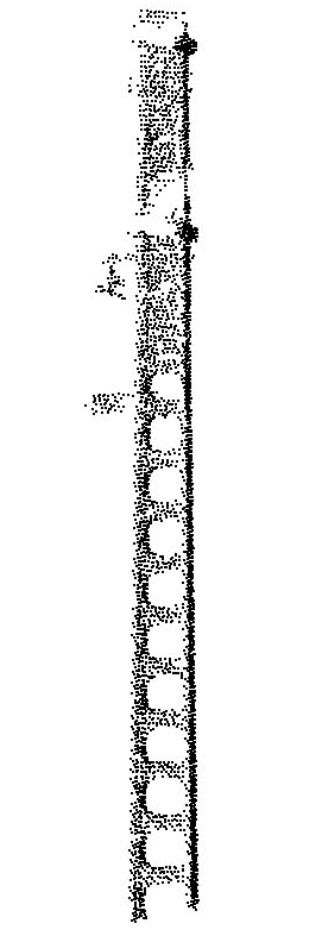}
  
\end{tabular}
\caption{Visual Demonstration of some examples from our dataset. For a video demonstration please refer to the video in the supplementary.}
\label{tab:cate_2}
\end{table*}

\begin{table*}
\centering
\setlength{\tabcolsep}{7pt}
\begin{tabular}{c c c c}
 \includegraphics[width=0.20\linewidth,trim=7cm 2cm 0cm 2cm,clip]{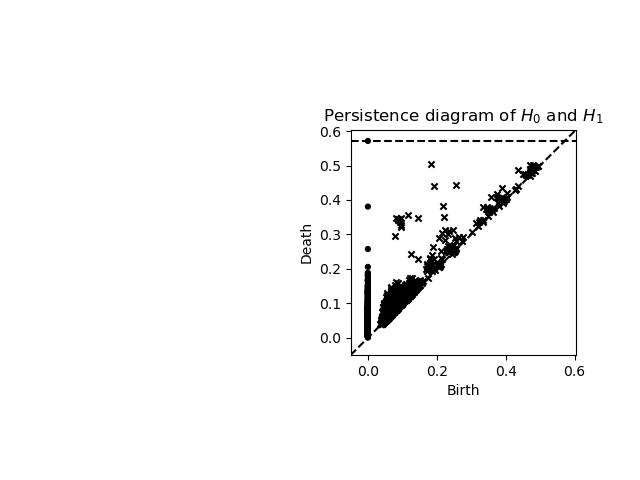}
 &\includegraphics[width=0.20\linewidth,trim=7cm 2cm 0cm 2cm,clip]{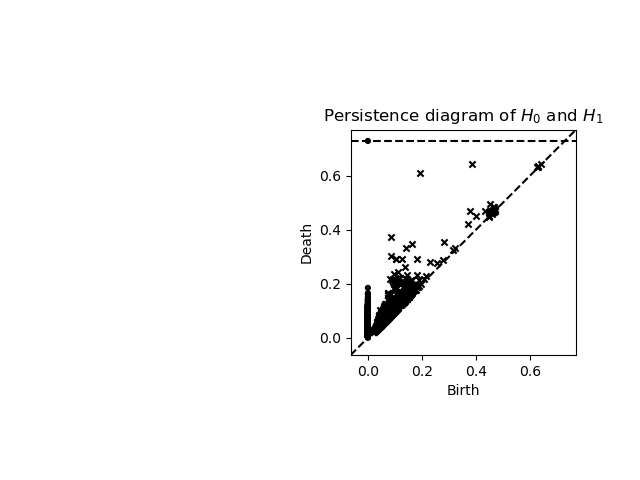}
 & \includegraphics[width=0.20\linewidth,trim=7cm 2cm 0cm 2cm,clip]{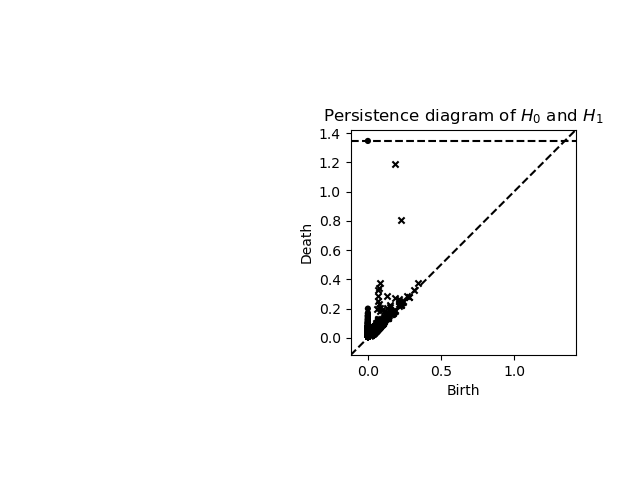}
 \\
\includegraphics[width=0.20\linewidth,trim=7cm 2cm 0cm 2cm,clip]{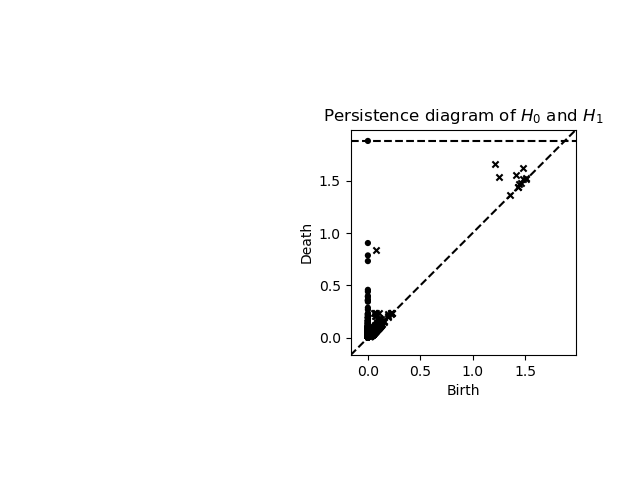}& \includegraphics[width=0.20\linewidth,trim=7cm 2cm 0cm 2cm,clip]{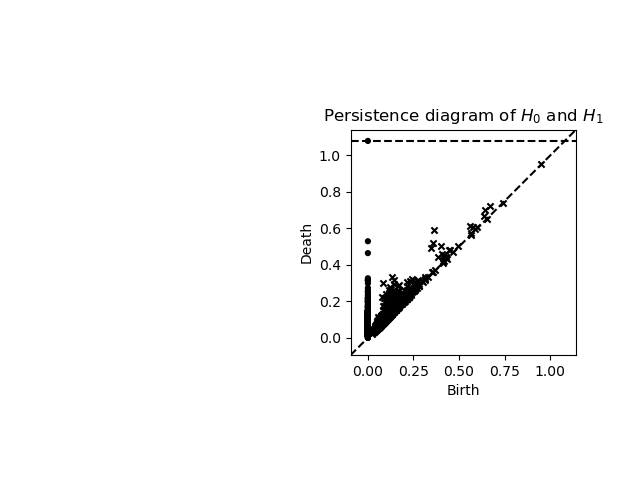}& \includegraphics[width=0.20\linewidth,trim=7cm 2cm 0cm 2cm,clip]{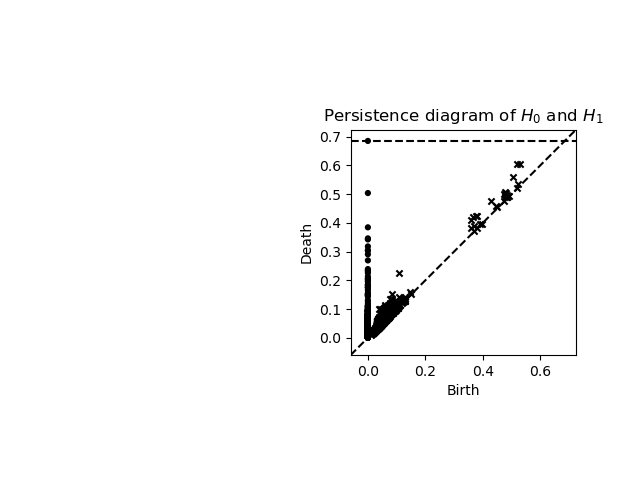}
\\

\includegraphics[width=0.20\linewidth,trim=7cm 2cm 0cm 2cm,clip]{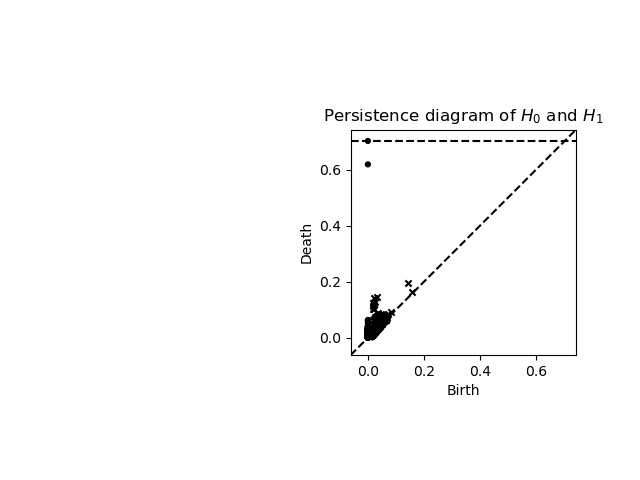}
& \includegraphics[width=0.20\linewidth,trim=7cm 2cm 0cm 2cm,clip]{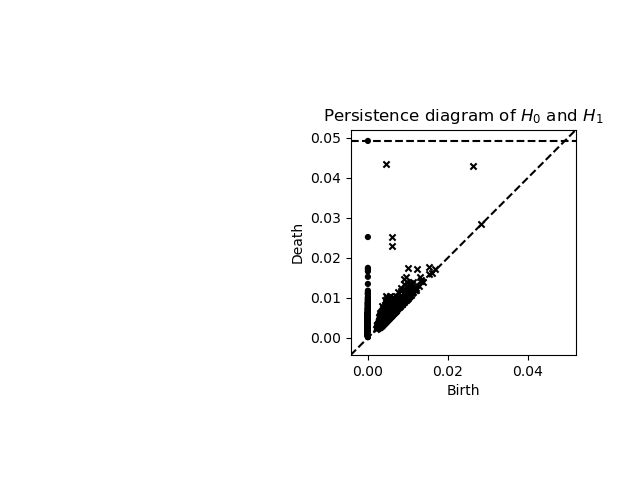}& \includegraphics[width=0.20\linewidth,trim=7cm 2cm 0cm 2cm,clip]{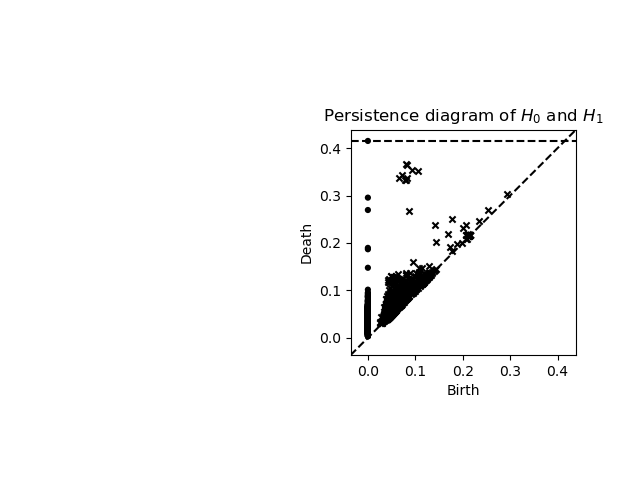}
\\
 
\end{tabular}
\caption{Persistence Diagram of nine point clouds (of Table \ref{tab:cate_1} and \ref{tab:cate_2}) of the \real{} dataset.}
\label{tab:pd}
\end{table*}

\begin{table*}
\centering
\setlength{\tabcolsep}{7pt}
\begin{tabular}{c c c c}
 \includegraphics[width=0.20\linewidth,trim=7cm 2cm 0cm 2cm,clip]{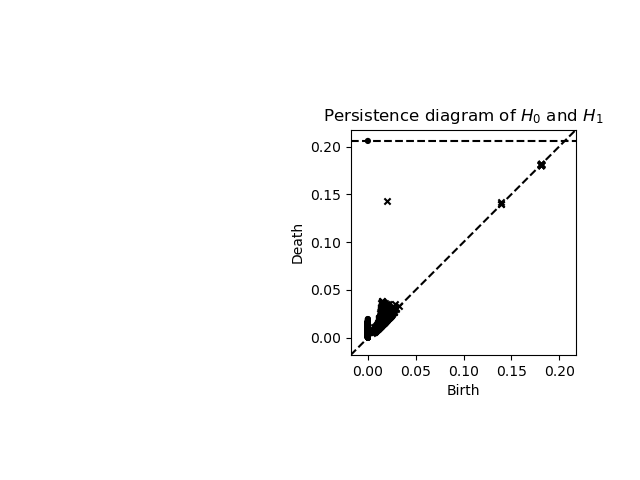}
 &\includegraphics[width=0.20\linewidth,trim=7cm 2cm 0cm 2cm,clip]{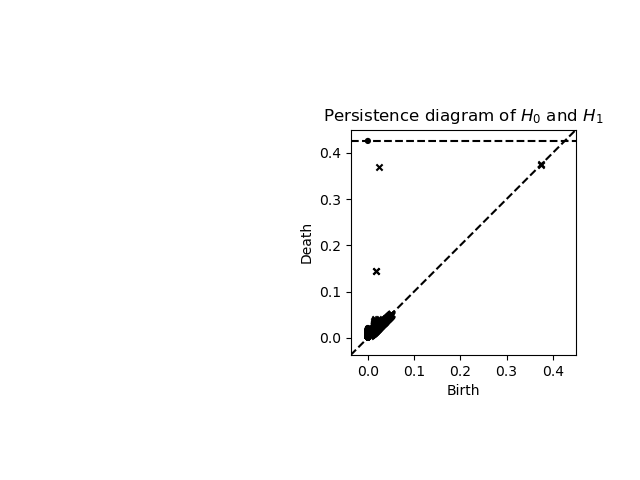}
 & \includegraphics[width=0.20\linewidth,trim=7cm 2cm 0cm 2cm,clip]{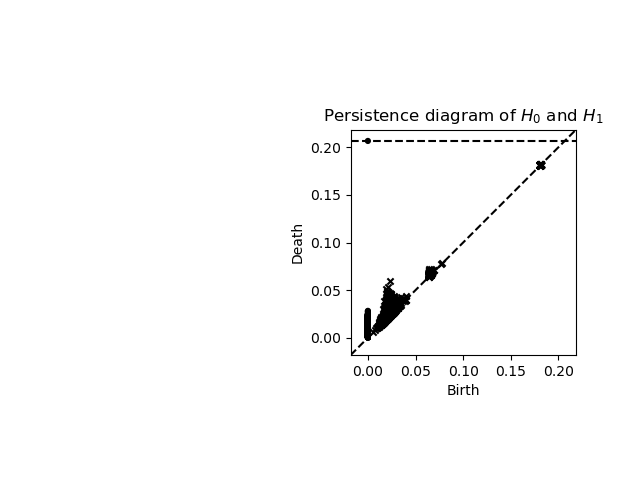}
 \\
\includegraphics[width=0.20\linewidth,trim=7cm 2cm 0cm 2cm,clip]{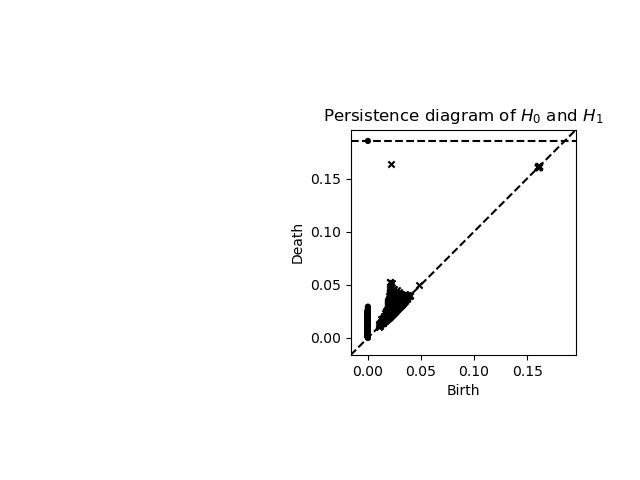}& \includegraphics[width=0.20\linewidth,trim=7cm 2cm 0cm 2cm,clip]{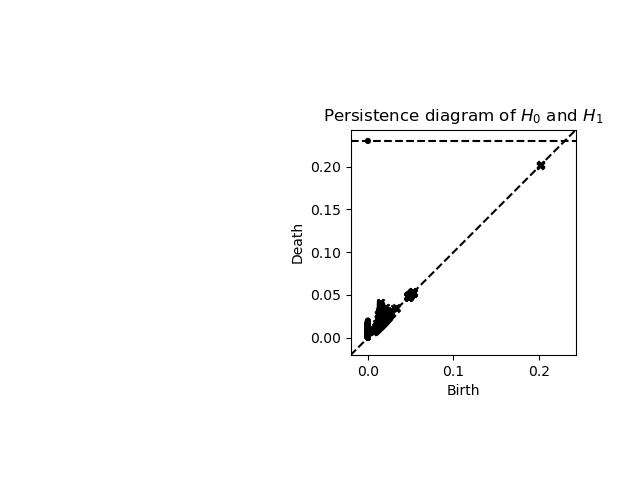}& \includegraphics[width=0.20\linewidth,trim=7cm 2cm 0cm 2cm,clip]{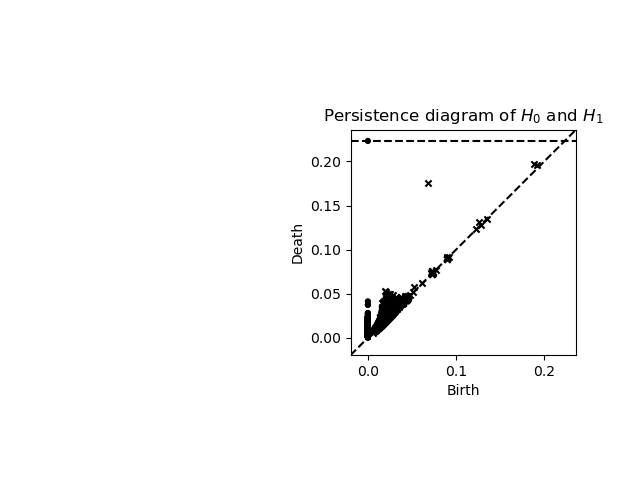}
\\

\includegraphics[width=0.20\linewidth,trim=7cm 2cm 0cm 2cm,clip]{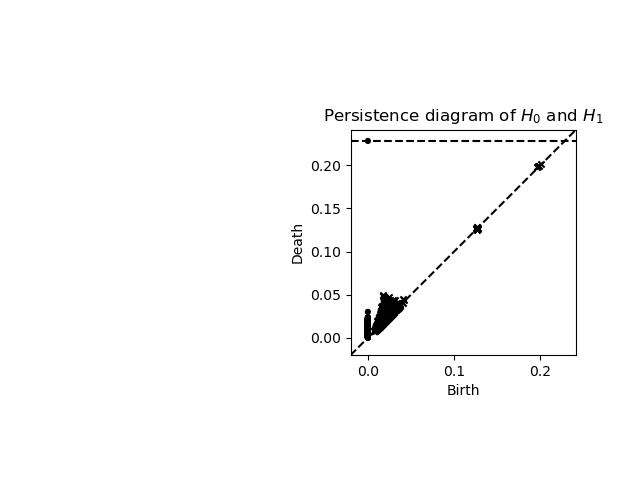}
& \includegraphics[width=0.20\linewidth,trim=7cm 2cm 0cm 2cm,clip]{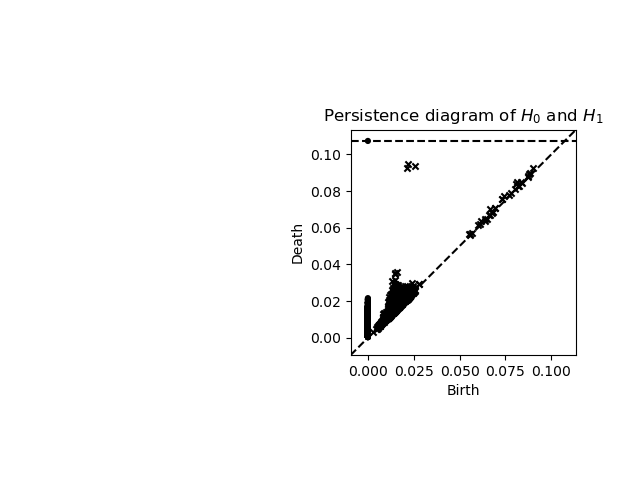}& \includegraphics[width=0.20\linewidth,trim=7cm 2cm 0cm 2cm,clip]{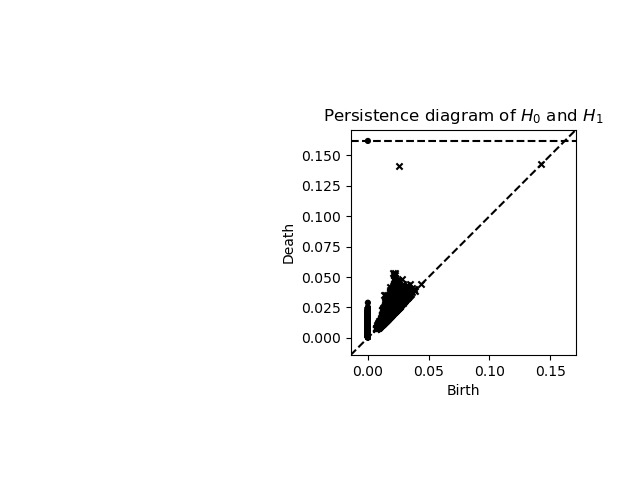}
\\
 
\end{tabular}
\caption{Persistence Diagram of nine point clouds from the synthetic datasets - ShapeNet and PCN.}
\label{tab:pd_shapenet}
\end{table*}


\end{document}